\documentclass[10pt,a4paper]{article}
\usepackage[utf8]{inputenc}
\usepackage[english]{babel}
\usepackage{url}
\usepackage{hyperref}
\usepackage{amsmath}
\usepackage{amsfonts}
\usepackage{amssymb}
\usepackage{graphicx}
\usepackage{float}
\usepackage{lipsum}
\usepackage{multicol}
\usepackage{epstopdf}
\usepackage{xcolor}
\usepackage{enumitem}
\usepackage{tabularx,booktabs,makecell}
\usepackage{algorithm}
\usepackage{cite}
\usepackage{graphicx}
\usepackage{amsmath,amssymb,amsfonts}
\usepackage{algpseudocode}
\usepackage{textcomp}
\usepackage{xcolor}
\usepackage{blindtext}
\usepackage{cuted}
\usepackage{physics}
\usepackage{xcolor}
\usepackage[toc,page]{appendix}
\usepackage[long]{optidef}
\usepackage[detect-none]{siunitx}
\usepackage{multirow}
\usepackage{array}

\definecolor{azul}{rgb}{0.0, 0.53, 0.74}
\usepackage{float}

\usepackage[left=2.00cm, right=2.00cm, top=2.00cm, bottom=2.00cm]{geometry}
\author{Aprendiendo \LaTeX\,}
\title{Mi primer paper}

\begin{document}
	\vspace{7.5mm}
	
	\begin{center}
		{\Large \textbf{Spatio–Temporal Graph Neural Networks for Dairy Farm Sustainability Forecasting and Counterfactual Policy Analysis}}\\
		\vspace{2mm}
		{\large Surya~Jayakumar, Kieran~Sullivan, John~McLaughlin, Christine O'Meara, and~Indrakshi Dey}\\
		\vspace{7.5mm}
		\textit{Walton Institute, South East Technological University, Waterford, Ireland} 
	\end{center}

\begin{abstract}
Efficiently steering dairy production toward environmental and operational sustainability requires models that jointly exploit spatial, temporal, and management information, rather than static indicator sets or purely local assessments. We propose a deterministic, data-driven framework that constructs an interpretable composite sustainability index and forecasts county-level sustainability trajectories using a Spatio–Temporal Graph Neural Network (STGNN) with temporal attention. County-level herd statistics from the Irish Cattle Breeding Federation (2021–2025) are first augmented via a Variational Autoencoder, expanding the dataset from 130 to 650 temporally consistent samples while preserving the joint distribution of 16 key operational indicators. Principal Component Analysis is then applied to identify four farm-level pillars : Reproductive Efficiency, Genetic Management, Herd Health, and Herd Management, and to derive a weighted composite sustainability index that serves as an operational proxy for environmental performance. The STGNN encodes spatial dependencies through an inter-county graph and captures non-linear temporal dynamics via attention-weighted histories, enabling multi-year sustainability forecasts for 2026–2030. The proposed model achieves high predictive skill (validation coefficient of determination above 0.91 with low mean absolute error) and generates smoother, more realistic trajectories than a Gaussian Kernel Regression baseline, particularly in sparsely sampled regions. Monte Carlo simulations quantify forecast uncertainty, while counterfactual scenario analyses for representative counties (like, Cork, Galway, and Kerry) illustrate how targeted interventions in fertility management, replacement rates, and culling strategies can shift long-term sustainability pathways. The framework links herd-level management decisions to regional sustainability outcomes and offers a transferable methodology for data-driven policy design in livestock systems.
\end{abstract}
\textbf{keywords} - Smart grid, 5G, URLLC, SDEN, Grid optimisation

\section{Introduction}\label{sec1}

Sustainability is now a defining requirement in dairy production, demanding that economic performance be balanced with environmental stewardship and animal welfare for long-term viability \cite{Mohammadzadeh2025}. However, sustainability is inherently dynamic and multi-factor, shaped by interacting biological processes, management choices, and external drivers that vary across space and time; therefore, indicator-by-indicator monitoring can be informative yet incomplete, often missing system-level trade-offs and spatio-temporal dependence \cite{Xu2023}.

In Ireland, dairy farming is vital to rural livelihoods and national output, but rising regulatory, climatic, and market pressures require robust, data-driven tools for farm- and policy-level decision-making \cite{AHI2024, deOliveira2024, Shine2022}. Herd-level data are structured into four operational pillars, \emph{Genetic Management}, \emph{Herd Health}, and \emph{Herd Management} \cite{Sommerseth2024,Crowe2018,Rios2020}, which operationalize the sustainability 3Ps (social, environmental, and economic) \cite{Diaz2021, Arvidsson2020}. Herd Health mainly supports the social pillar via welfare, while the remaining pillars primarily drive economic performance with indirect environmental gains through improved efficiency \cite{VanEenennaam2025}. This pillar-based formulation links actionable improvements to environmental outcomes. Better fertility (e.g., shorter calving intervals, higher conception rates, and lower age at first calving) reduces replacement and other nonproductive animals \cite{Diavao2023}, lowering feed demand, manure output, and methane emissions. Genetic improvement and stronger health/management practices further increase output per unit input and reduce losses and involuntary culling, indirectly conserving land, water, and energy \cite{Crowe2018,VanEenennaam2025}. By quantifying these effects in a unified scoring framework, the approach enables consistent inter-farm and inter-regional benchmarking, supports identification of high-performing reference systems, and highlights priority targets for intervention \cite{deHaas2023}.

This paper presents a \emph{novel} spatio-temporal, data-driven framework to assess, forecast, and stress-test dairy sustainability across Irish counties. To the best of the knowledge, this is the \emph{first-ever} county-scale framework that jointly (i) derives an interpretable composite sustainability score from herd-level operational records and (ii) predicts its evolution using spatio-temporal graph neural networks (STGNNs) that explicitly model both geographic interactions and temporal dynamics\cite{Muruganantham2022, Gupta2023, Vyas2020, Sophocleous2024}. Using Irish Cattle Breeding Federation (ICBF) data \cite{ICBF2025}, the proposed pipeline integrates: synthetic data generation to mitigate sparsity and support scenario exploration; identification of operational sustainability pillars; construction of composite sustainability scores; and STGNN-based forecasting for forward-looking assessment. The same modeling backbone is further used to conduct counterfactual (what-if) analyses that quantify how plausible changes in management-relevant factors can shift future sustainability outcomes. The main contributions of this work are as follows:

\begin{itemize}
    \item Introduce a \textbf{novel, end-to-end} spatio-temporal pipeline that converts herd-level operational data into county-level composite sustainability scores and trajectories suitable for regional benchmarking.
    \item Propose an \textbf{interpretable pillar-based scoring formulation} built on four operational pillars (Reproductive Efficiency, Genetic Management, Herd Health, and Herd Management) that enables consistent comparison across counties and years.
    \item Develop the \textbf{first-ever county-scale STGNN forecasting setup} for dairy sustainability that captures geographic dependence and non-stationary temporal dynamics within a unified predictive model.
    \item Enable \textbf{counterfactual (what-if) scenario analysis} to quantify how plausible changes in operational levers could alter future sustainability outcomes, supporting actionable and evidence-based decision-making.
\end{itemize}

\section{Methodology for Data Pipeline}

\begin{figure*}[t!]
\centering
\includegraphics[width=1\textwidth]{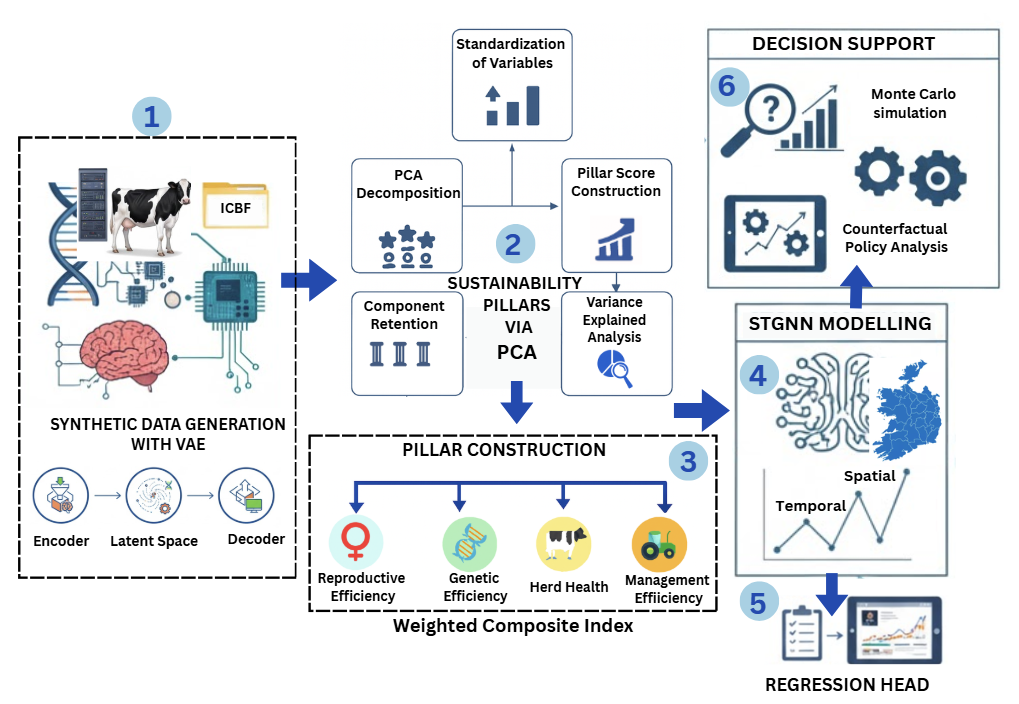}
\vspace{-3mm}
\caption{Concept diagram of the entire framework}
\vspace{-5mm}
\label{fig:framework}
\end{figure*}

The proposed framework for assessing dairy farm sustainability follows a structured pipeline that integrates data preprocessing, composite sustainability scoring, STGNNs, scenario simulations, and temporal visualization (Fig. \ref{fig:framework}) First, county-level dairy statistics are cleaned, filtered, and normalized to create a consistent input space suitable for deep learning. Synthetic yet statistically consistent samples are then generated to enrich the temporal depth of the dataset. Next, a composite sustainability index is constructed from key operational pillars using Principal Component Analysis (PCA), providing an interpretable target variable that aggregates reproductive, genetic, health, and management dimensions. This index is subsequently modeled using an STGNN that explicitly captures spatial dependencies between counties and temporal dynamics across years. Finally, Monte Carlo–based scenario simulations and temporal visualization tools are employed to explore uncertainty, evaluate counterfactual policy interventions, and illustrate how sustainability trajectories evolve over the forecasting horizon.

\subsection{Dataset Description and Preprocessing}
\label{subsec:data}

County-level dairy statistics from the Irish Cattle Breeding Federation (ICBF) for 2021–2025 were used \cite{ICBF2025}. Let $i\in\{1,\dots,N\}$ index counties ($N=26$) and $t\in\{2021,\dots,2025\}$ index years. For each $(i,t)$ pair,  an $m$-dimensional vector of herd-level operational indicators $X_{i,t}\in\mathbb{R}^{m}$ describing herd demographics, reproductive efficiency, replacement dynamics, and animal health was observed. Key variables include the number of herds, calving interval, 6-week calving rate (including seasonal variants), calves per cow per year, average lactations, current and potential replacement rates, culling rate, heifers calved at 22-26 months, and mortality rates (at birth and at 28 days). Breeding and data-quality indicators such as replacements bred to dairy AI, births with known sire, and births with calving survey data are also included, alongside recycled cows and autumn calving statistics. Although direct measurements of greenhouse gas emissions or nutrient losses are not used, these indicators are acted upon as operational proxies for environmental performance, improved fertility (e.g., shorter calving interval and higher early calving rates), lower involuntary culling, and lower young-stock mortality typically increase lifetime productivity and reduce resource use per unit output, thereby lowering emissions intensity. Hence, aggregating $X_{i,t}$ into a composite sustainability index captures management-driven variation that is tightly coupled to environmental footprint, even without explicit emissions inventories.

Most indicators originate from on-farm digital workflows rather than manual reporting. Animals are uniquely tracked via electronic identification (EID), while events such as births, deaths, calvings, and movements are recorded through farm software and handheld devices and integrated with national identification and movement systems. In addition, milking systems and on-animal sensors (e.g., activity and health monitors) support automated capture of reproduction- and health-relevant signals that feed into herd management decisions and subsequent county-year summaries in the ICBF database. As a result, the dataset reflects information streams commonly available in modern dairy operations. Prior to modeling, a standard preprocessing pipeline ws applied. First, records for completeness and internal consistency across counties and years, removing structurally impossible values according to ICBF definitions were screened. Remaining missing entries are handled with continuity-preserving cleaning to maintain each county’s time series. Next, a regular spatio-temporal panel was constructd by sorting all samples by $(i,t)$. Finally, each feature was scaled using Min-Max normalization to obtain $x^{(j)}_{i,t}\in[0,1]$, preventing large-magnitude variables (e.g., number of herds) from dominating optimization and ensuring that reproductive, genetic, health, and management indicators contribute comparably to composite scoring and STGNN training.

\subsection{Synthetic Data Generation with Variational Autoencoder (VAE)}
\label{subsec:vae}
The original panel spans only five years (2021-2025), which limits the temporal diversity available for training and can reduce robustness in multi-year forecasting. To mitigate data sparsity,  the county-year samples were augmented using a variational autoencoder (VAE) that learns a low-dimensional probabilistic representation of the normalized feature vectors and generates additional synthetic observations that preserve the multivariate dependence structure of the herd-level indicators. Let $\mathbf{x}\in\mathbb{R}^{D}$ denote a normalized county-year feature vector, where $D$ is the number of indicators in Section~\ref{subsec:data}. The VAE consists of an encoder $q_{\phi}(\mathbf{z}\mid\mathbf{x})$ and a decoder $p_{\theta}(\mathbf{x}\mid\mathbf{z})$, parameterized by $\phi$ and $\theta$, respectively. The encoder maps $\mathbf{x}$ to a latent variable $\mathbf{z}\in\mathbb{R}^{L}$ (latent dimension $L=80$ in the implementation used here), while the decoder reconstructs $\mathbf{x}$ from $\mathbf{z}$.

\subsubsection{Encoder and Reparameterization}

The encoder outputs a mean vector $\boldsymbol{\mu}(\mathbf{x})\in\mathbb{R}^{L}$ and a log-variance vector $\log\boldsymbol{\sigma}^2(\mathbf{x})\in\mathbb{R}^{L}$, defining a diagonal Gaussian approximate posterior,
\begin{equation}
q_{\phi}(\mathbf{z}\mid\mathbf{x})=\mathcal{N}\!\left(\mathbf{z};\,\boldsymbol{\mu}(\mathbf{x}),\,\mathrm{diag}\big(\boldsymbol{\sigma}^2(\mathbf{x})\big)\right),
\end{equation}
where $\boldsymbol{\sigma}^2(\mathbf{x})=\exp(\log\boldsymbol{\sigma}^2(\mathbf{x}))$ and $\mathrm{diag}(\cdot)$ forms a diagonal covariance matrix. To enable gradient-based optimization, $\mathbf{z}$ is sampled via the reparameterization trick.
\begin{equation}
\mathbf{z}=\boldsymbol{\mu}(\mathbf{x})+\boldsymbol{\sigma}(\mathbf{x})\odot\boldsymbol{\epsilon}, \quad
\boldsymbol{\epsilon}\sim\mathcal{N}(\mathbf{0},\mathbf{I}_{L}),
\label{eq:reparam}
\end{equation}
where $\boldsymbol{\sigma}(\mathbf{x})=\sqrt{\boldsymbol{\sigma}^2(\mathbf{x})}$, $\odot$ denotes element-wise multiplication, and $\mathbf{I}_{L}$ is the $L\times L$ identity matrix. Sampling from $p_{\theta}(\mathbf{x}\mid\mathbf{z})$ then yields synthetic county-year feature vectors that expand the effective training set while remaining statistically consistent with the observed data.

\subsubsection{Decoder and Reconstruction}

The decoder $p_{\theta}(\mathbf{x}\mid\mathbf{z})$ mirrors the encoder with fully connected layers of increasing width and ReLU activations, followed by a linear output layer. Given a latent vector $\mathbf{z}$, the decoder produces a reconstruction $\hat{\mathbf{x}} \in \mathbb{R}^{D}$ according to $\hat{\mathbf{x}} = f_{\theta}(\mathbf{z}),$ where $f_{\theta}(\cdot)$ represents the deterministic mapping implemented by the decoder network. Under a Gaussian observation model with fixed diagonal covariance, minimizing the mean squared error between $\mathbf{x}$ and $\hat{\mathbf{x}}$ is equivalent to maximizing the log-likelihood of $\mathbf{x}$ given $\mathbf{z}$.

\subsubsection{Training Objective}
Let $\{\mathbf{x}_n\}_{n=1}^{N}$ be the set of $N$ normalized county-year feature vectors, with $\mathbf{x}_n\in\mathbb{R}^{D}$. A $\beta$-VAE was trained by minimizing the average reconstruction–regularization objective.
\begin{equation}
\mathcal{L}(\theta,\phi)=\frac{1}{N}\sum_{n=1}^{N}\Big(\|\mathbf{x}_n-\hat{\mathbf{x}}_n\|_2^2+\beta\,\mathrm{KL}\big(q_{\phi}(\mathbf{z}\mid\mathbf{x}_n)\,\|\,p(\mathbf{z})\big)\Big),
\label{eq:vae_loss}
\end{equation}
where $\hat{\mathbf{x}}_n=f_{\theta}(\mathbf{z}_n)$ is the decoder reconstruction, $\|\cdot\|_2$ is the Euclidean norm, $\mathrm{KL}(\cdot\|\cdot)$ is the Kullback-Leibler divergence, $p(\mathbf{z})=\mathcal{N}(\mathbf{0},\mathbf{I}_{L})$ is the latent prior, $L$ is the latent dimension, and $\beta>0$ controls the regularization strength. Since $q_{\phi}(\mathbf{z}\mid\mathbf{x})=\mathcal{N}(\boldsymbol{\mu}(\mathbf{x}),\mathrm{diag}(\boldsymbol{\sigma}^2(\mathbf{x})))$, the KL term has the closed form
\begin{equation}
\mathrm{KL}\big(q_{\phi}(\mathbf{z}\mid\mathbf{x})\,\|\,p(\mathbf{z})\big)
=\frac{1}{2}\sum_{j=1}^{L}\Big(\mu_j(\mathbf{x})^2+\sigma_j(\mathbf{x})^2-\log\sigma_j(\mathbf{x})^2-1\Big),
\label{eq:kl_closed_form}
\end{equation}
where $\mu_j(\mathbf{x})$ and $\sigma_j(\mathbf{x})$ are the $j$th components of $\boldsymbol{\mu}(\mathbf{x})$ and $\boldsymbol{\sigma}(\mathbf{x})$, respectively.

\subsubsection{Synthetic Sample Generation and Dataset Augmentation}
After training, synthetic county-year feature vectors were generated to expand the effective sample size for downstream spatio-temporal modeling. In unconditional sampling, $\mathbf{z}{\mathrm{new}}\sim\mathcal{N}(\mathbf{0},\mathbf{I}{L})$ was drawn and decoded as $\mathbf{x}{\mathrm{syn}} = f{\theta}(\mathbf{z}_{\mathrm{new}})$, producing samples consistent with the globally learned distribution. In conditional augmentation, the neighborhood was enriched around each observed instance $\mathbf{x}_n$ by sampling $K$ latent perturbations from its approximate posterior. Specifically, for $k\in\{1,\dots,K\}$ $\boldsymbol{\epsilon}^{(k)}\sim\mathcal{N}(\mathbf{0},\mathbf{I}_{L})$ was sampled and formed.
\begin{equation}
\mathbf{z}^{(k)}_n=\boldsymbol{\mu}(\mathbf{x}_n)+\boldsymbol{\sigma}(\mathbf{x}_n)\odot \boldsymbol{\epsilon}^{(k)}, 
\qquad
\tilde{\mathbf{x}}^{(k)}_n=f_{\theta}(\mathbf{z}^{(k)}_n),
\end{equation}
where $\odot$ denotes element-wise multiplication and $K$ is the number of synthetic replicates per real sample. This strategy increases the number of county-year vectors available for training while preserving the marginal ranges and cross-indicator dependencies captured from the original ICBF data.

\subsubsection{Robustness and Validation of Synthetic Data}
VAE-generated samples were validated using (i) distributional plausibility and (ii) downstream predictive utility. Let $\mathcal{X}_{\mathrm{real}}=\{\mathbf{x}_n\}_{n=1}^{N}$ denote the $N$ real county-year vectors and $\mathcal{X}_{\mathrm{syn}}=\{\mathbf{x}^{\mathrm{syn}}_m\}_{m=1}^{M}$ denote the $M$ synthetic vectors, where each $\mathbf{x}\in\mathbb{R}^{D}$ and $x_{n,d}$ is the $d$th feature component. For each feature $d\in\{1,\dots,D\}$,  empirical moments were compared via the real and synthetic means $\mu^{\mathrm{real}}_d=\frac{1}{N}\sum_{n=1}^{N}x_{n,d}$ and $\mu^{\mathrm{syn}}_d=\frac{1}{M}\sum_{m=1}^{M}x^{\mathrm{syn}}_{m,d}$ (and similarly variances $\sigma^{2,\mathrm{real}}_d$ and $\sigma^{2,\mathrm{syn}}_d$) to detect systematic bias.

To verify that cross-indicator dependence is preserved,  Pearson correlation matrices was computed $\mathbf{R}^{\mathrm{real}},\mathbf{R}^{\mathrm{syn}}\in\mathbb{R}^{D\times D}$ and quantified their discrepancy by the Frobenius norm
$\Delta_{\mathrm{corr}}=\|\mathbf{R}^{\mathrm{real}}-\mathbf{R}^{\mathrm{syn}}\|_{F}$,
where $\|\mathbf{A}\|_{F}=\sqrt{\sum_{i}\sum_{j}A_{ij}^{2}}$. A non-parametric distributional distance was also evaluated using the squared Maximum Mean Discrepancy (MMD). Given a positive-definite kernel $k(\cdot,\cdot)$ on $\mathbb{R}^{D}$ (e.g., a Gaussian kernel), the empirical MMD is
\begin{align}
&\mathrm{MMD}^{2}(\mathcal{X}_{\mathrm{real}},\mathcal{X}_{\mathrm{syn}})
=\frac{1}{N^{2}}\sum_{n=1}^{N}\sum_{n'=1}^{N}k(\mathbf{x}_n,\mathbf{x}_{n'}) \nonumber\\
&+ \frac{1}{M^{2}}\sum_{m=1}^{M}\sum_{m'=1}^{M}k(\mathbf{x}^{\mathrm{syn}}_m,\mathbf{x}^{\mathrm{syn}}_{m'})-\frac{2}{NM}\sum_{n=1}^{N}\sum_{m=1}^{M}k(\mathbf{x}_n,\mathbf{x}^{\mathrm{syn}}_m),
\end{align}
where values near $0$ indicate close alignment between real and synthetic distributions in the kernel-induced feature space.

Finally, predictive utility was tested by training the downstream STGNN under two settings using identical architecture and hyperparameters: (i) real-only training set $\mathcal{X}_{\mathrm{real}}$, and (ii) augmented training set $\mathcal{X}_{\mathrm{real}}\cup\mathcal{X}_{\mathrm{syn}}$. Both models are evaluated on the same held-out \emph{real} test set $\{(y_t,\hat{y}_t)\}_{t=1}^{T}$, where $y_t$ is the true sustainability score, $\hat{y}_t$ is the prediction, and $T$ is the number of test instances. $\mathrm{MAE}=\frac{1}{T}\sum_{t=1}^{T}|y_t-\hat{y}_t|$, $\mathrm{RMSE}=\sqrt{\frac{1}{T}\sum_{t=1}^{T}(y_t-\hat{y}_t)^2}$, and $R^{2}=1-\frac{\sum_{t}(y_t-\hat{y}_t)^2}{\sum_{t}(y_t-\bar{y})^2}$ with $\bar{y}=\frac{1}{T}\sum_{t=1}^{T}y_t$ were reported. Comparable or improved $R^{2}$ with reduced MAE/RMSE under augmentation indicates that synthetic samples preserve statistical realism and add useful variability for robust forecasting.

\subsection{Sustainability Pillar Identification using PCA}
\label{subsec:pca}
To identify latent sustainability pillars from operational indicators,  Principal Component Analysis (PCA) was applied to $p=16$ standardized variables describing reproduction, replacement dynamics, herd health, and management (Section~\ref{subsec:data}). Let $X\in\mathbb{R}^{N\times p}$ denote the county-year data matrix, where $N$ is the number of observations and each row is one county-year sample. Each column was standardized to zero mean and unit variance to obtain $X^{\mathrm{sc}}\in\mathbb{R}^{N\times p}$, ensuring that all indicators contribute comparably.

PCA is performed on the sample covariance matrix
$S=\frac{1}{N-1}(X^{\mathrm{sc}})^{\top}X^{\mathrm{sc}}\in\mathbb{R}^{p\times p}$.
Eigendecomposition was computed $S=V\Lambda V^{\top}$, where $V\in\mathbb{R}^{p\times p}$ contains orthonormal eigenvectors and $\Lambda=\mathrm{diag}(\lambda_1,\dots,\lambda_p)$ contains eigenvalues ordered as $\lambda_1\geq\dots\geq\lambda_p\geq 0$. The leading eigenvectors define directions of maximum variance and the corresponding $\lambda_k$ quantify explained variance.

$K=4$ components are retained to obtain interpretable farm-level pillars while capturing the dominant variability. Let $W=[\mathbf{v}_1,\dots,\mathbf{v}_K]\in\mathbb{R}^{p\times K}$ be the loading matrix formed by the first $K$ eigenvectors. The pillar (component) scores are
\begin{equation}
Z=X^{\mathrm{sc}}W\in\mathbb{R}^{N\times K},
\end{equation}
where $z_{n,k}$ (entry $(n,k)$ of $Z$) is the $k$th pillar score for observation $n$, computed as the linear combination
$z_{n,k}=\sum_{j=1}^{p}x^{\mathrm{sc}}_{n,j}W_{j,k}$,
with $x^{\mathrm{sc}}_{n,j}$ the standardized value of feature $j$ and $W_{j,k}$ its loading on pillar $k$.

The variance explained by pillar $k$ is quantified by the variance ratio
\begin{equation}
\rho_k=\frac{\lambda_k}{\sum_{i=1}^{p}\lambda_i},
\end{equation}
and the cumulative explained variance is $\sum_{k=1}^{K}\rho_k$. In the application used , the first four components capture the majority of variance, while additional components provide marginal gains and reduced interpretability, motivating $K=4$ for downstream scoring and forecasting.

\subsubsection{Linking PCA Components to Sustainability Pillars}
 the retained PCA components were mapped to domain-interpretable sustainability pillars using a simple loadings-based procedure with sign calibration. Let $W\in\mathbb{R}^{p\times K}$ denote the PCA loading matrix (Section~\ref{subsec:pca}), where $W_{j,k}$ is the loading of feature $j\in\{1,\dots,p\}$ on component $k\in\{1,\dots,K\}$ and $K=4$. Feature influence on component $k$ is quantified by the absolute loading $\ell_{j,k}=|W_{j,k}|$. The set of dominant features were defined  for component $k$ as $\mathcal{J}_k=\{j:\ell_{j,k}\ge \tau\}$, where $\tau$ is a loading threshold (chosen near the 75th percentile of $\{\ell_{j,k}\}_{j=1}^{p}$) to retain a small set of clearly influential indicators.

Then component $k$ was assigned to a pillar based on the thematic content of $\mathcal{J}_k$ (e.g., fertility- and calving-related features imply \emph{Reproductive Efficiency}; mortality and culling imply \emph{Herd Health}; breeding/sire and AI indicators imply \emph{Genetic Management}; replacement structure and heifer age indicators imply \emph{Herd Management}). Because PCA directions are sign-indeterminate, a consistent ``higher-is-better'' interpretation was enforced by introducing an orientation factor $s_k\in\{+1,-1\}$ and defining reoriented loadings and scores as $\tilde{W}_{j,k}=s_k W_{j,k}$ and $\tilde{z}_{n,k}=s_k z_{n,k}$, where $z_{n,k}$ is the original component score for observation $n$ (Section~\ref{subsec:pca}). $s_k$ was chosen such that increases in $\tilde{z}_{n,k}$ corresponded to improved performance in the associated pillar.

\subsubsection{Environmental Interpretation of Pillars}
Although direct emissions are not observed, the pillars have a mechanistic link to environmental efficiency. Let $s_{n,k}$ denote the normalized, reoriented pillar score for observation $n$ and pillar $k$. Higher $s_{n,k}$ for \emph{Reproductive Efficiency} corresponds to improved fertility and more compact calving, reducing nonproductive days and replacement pressure, which lowers resource use and emissions intensity per unit output. Higher $s_{n,k}$ for \emph{Herd Health} reflects lower mortality and involuntary culling, reducing the number of replacement animals required and the associated rearing burden. \emph{Herd Management} captures replacement rates and age structure; higher scores indicate more efficient herd turnover and better alignment of inputs (feed/land) with productive output. \emph{Genetic Management} reflects breeding strategy; higher scores indicate selection and data quality that support persistence and productivity over multiple lactations, reducing inputs per unit output over time. After reorientation, pillar scores were stanadized across all $N$ observations to obtain $S\in\mathbb{R}^{N\times K}$ with entries $s_{n,k}$ (zero mean and unit variance per pillar). These $K=4$ normalized pillars are subsequently used to construct the composite sustainability index and to train the spatio-temporal forecasting models.

\subsubsection{Robustness of Pillar Structure}
Pillar robustness was assessed in three ways. First, the eigenvalue spectrum $\{\lambda_k\}_{k=1}^{p}$ is inspected via scree and cumulative variance to justify $K=4$; additional components provide diminishing variance gains and do not alter the interpretation of the first four pillars. Second, loading stability is evaluated via bootstrap resampling: for each bootstrap replicate $b$, PCA yields loadings $W^{(b)}$, and the correlation between $W_{\cdot,k}$ and $W^{(b)}_{\cdot,k}$ was computed; consistently high correlations indicate stable feature importance and stable $\mathcal{J}_k$. Third, PCA was repeated under alternative scaling choices and compare loading vectors via cosine similarity, confirming that the dominant groupings remain consistent.

Finally, relevance to sustainability outcomes was validated by correlating pillar scores with the composite sustainability index. Let $y_n$ be the composite index for observation $n$, and let $\bar{s}_k$ and $\bar{y}$ be sample means. The Pearson correlation between pillar $k$ and the index is
\begin{equation}
\rho_{k,y}=\frac{\sum_{n=1}^{N}(s_{n,k}-\bar{s}_k)(y_n-\bar{y})}
{\sqrt{\sum_{n=1}^{N}(s_{n,k}-\bar{s}_k)^2}\sqrt{\sum_{n=1}^{N}(y_n-\bar{y})^2}}.
\end{equation}
Pillars tied to fertility and loss reduction (e.g., \emph{Reproductive Efficiency} and \emph{Herd Health}) exhibit strong positive association with $y_n$ and negative association with undesirable indicators such as long calving intervals or high replacement rates, supporting the interpretation that the PCA structure captures sustainability-relevant operational variation.

\subsection{Weighted Pillar-Based Sustainability Scoring Framework}
\label{subsec:score}
After identifying $K=4$ PCA-based pillars (Section~\ref{subsec:pca}), a county-year sustainability score was contructed by (i) forming weighted pillar scores from dominant indicators and (ii) aggregating pillars into a single index. Let $X^{\mathrm{sc}}\in\mathbb{R}^{N\times p}$ be the standardized feature matrix ($p=16$ indicators, $N$ county-year observations), and let $W\in\mathbb{R}^{p\times K}$ be the PCA loading matrix. Because PCA signs are arbitrary,  an orientation factor $s_k\in\{+1,-1\}$ ws applied per pillar to enforce a higher-is-better direction, yielding oriented loadings $\tilde{W}_{j,k}=s_k W_{j,k}$.

For each pillar $k$, a compact set of dominant indicators
$\mathcal{J}k={j\in{1,\dots,p}:,|\tilde{W}{j,k}|\ge \tau}$ was retained,
where $\tau>0$ was a loading threshold (chosen near the upper quartile of $|\tilde{W}_{\cdot,k}|$). Nonnegative within-pillar weights were then defined by normalizing the absolute oriented loadings:
\begin{equation}
w_{j,k}=
\begin{cases}
\displaystyle\frac{|\tilde{W}_{j,k}|}{\sum_{\ell\in\mathcal{J}_k}|\tilde{W}_{\ell,k}|}, & j\in\mathcal{J}_k,\\[8pt]
0, & j\notin\mathcal{J}_k,
\end{cases}
\label{eq:feature_weight}
\end{equation}
so that $w_{j,k}\ge 0$ and $\sum_{j=1}^{p}w_{j,k}=1$ for each pillar $k$.

Let $x^{\mathrm{sc}}_{n,j}$ denote the standardized value of feature $j$ for observation $n$. The raw pillar score for observation $n$ and pillar $k$ is the convex combination
\begin{equation}
P_{n,k}=\sum_{j\in\mathcal{J}_k} w_{j,k}\,x^{\mathrm{sc}}_{n,j},
\label{eq:pillar_score}
\end{equation}
and  pillar scores were standardized across the dataset to obtain
$S_{n,k}=(P_{n,k}-\bar{P}_k)/\sigma_{P_k}$,
where $\bar{P}_k=\frac{1}{N}\sum_{n=1}^{N}P_{n,k}$ and
$\sigma_{P_k}$ is the sample standard deviation of $\{P_{n,k}\}_{n=1}^{N}$.
The composite sustainability index for observation $n$ is then the equal-weight average
\begin{equation}
I_n=\frac{1}{K}\sum_{k=1}^{K}S_{n,k}, \quad K=4,
\label{eq:raw_index}
\end{equation}
which treats pillars symmetrically (consistent with their orthogonality and distinct variance contributions in PCA).

For interpretability,  $\{I_n\}_{n=1}^{N}$ was rescaled to $[0,100]$. Let
$I_{\min}=\min_n I_n$ and $I_{\max}=\max_n I_n$; the final sustainability score is
\begin{equation}
I_n^{\ast}=100\,\frac{I_n-I_{\min}}{I_{\max}-I_{\min}}.
\label{eq:scaled_index}
\end{equation}
Local sensitivity to a standardized indicator $x^{\mathrm{sc}}_{n,j}$ follows from
\eqref{eq:pillar_score}-\eqref{eq:raw_index}:
\begin{equation}
\frac{\partial I_n}{\partial x^{\mathrm{sc}}_{n,j}}
=\frac{1}{K}\sum_{k=1}^{K}\frac{\partial S_{n,k}}{\partial x^{\mathrm{sc}}_{n,j}}
\approx \frac{1}{K}\sum_{k:\, j\in\mathcal{J}_k}\frac{w_{j,k}}{\sigma_{P_k}},
\end{equation}
where the approximation neglects the weak dependence of $\bar{P}_k$ and $\sigma_{P_k}$ on a single observation. Hence, indicators with larger $w_{j,k}$ exert stronger influence, and the orientation embedded in $\tilde{W}_{j,k}$ ensures that increases in beneficial indicators improve the resulting sustainability score.

\subsubsection{Robustness and Validation of the Scoring Framework}
robustness of the scoring framework was assessed along three axes: (i) sensitivity to weighting, (ii) temporal/spatial stability, and (iii) alignment with operational sustainability proxies.

\textbf{Weight sensitivity:} The proposed within-pillar PCA weights $w_{j,k}$ (Section~\ref{subsec:score}) were compared against two alternatives. First, an equal-weight baseline $w^{\mathrm{eq}}_{j,k}=1/|\mathcal{J}_k|$ for $j\in\mathcal{J}_k$ (and $0$ otherwise), where $|\mathcal{J}_k|$ is the number of selected indicators in pillar $k$. Second, an eigenvalue-weighted composite that emphasizes pillars explaining more variance:
$I_n^{\lambda}=\sum_{k=1}^{K}\omega_k S_{n,k}$, where $\omega_k=\lambda_k/\sum_{i=1}^{K}\lambda_i$ and $\{\lambda_k\}_{k=1}^{K}$ are the PCA eigenvalues. For each variant, rescaled scores were computed (as in \eqref{eq:scaled_index}) and measured agreement with the proposed index $I_n^{\ast}$ using Spearman’s $\rho$ and Kendall’s $\tau$ rank correlations over all county-year observations $n$. High rank correlations indicate that county benchmarking is not an artifact of a particular weighting choice.

\textbf{Temporal and spatial stability:} For each county $c$, score stability across the available years was summarized by the mean $\bar{I}c$ and standard deviation $\sigma{I_c}$ of ${I_n^{\ast}}$ for that county. Counties with high $\bar{I}_c$ and low $\sigma_{I_c}$ exhibit consistently strong performance, whereas low $\bar{I}_c$ and/or high $\sigma_{I_c}$ indicate persistent gaps or instability. Additionally year-by-year rank correlations were computed across counties to verify that rankings are not dominated by transient fluctuations in individual indicators.

\textbf{Alignment with sustainability proxies:} The relationship between the composite index and pillar scores was quantified using the linear model, $I_n=\alpha_0+\sum_{k=1}^{K}\alpha_k S_{n,k}+\varepsilon_n$, where $\alpha_0$ is an intercept, $\alpha_k$ are coefficients, and $\varepsilon_n$ is a zero-mean residual. Because PCA pillars are orthogonal, coefficient interpretation is stable and reflects each pillar’s contribution to index variability. $I_n^{\ast}$ was further regressed against key operational proxies (e.g., calving interval and replacement rate) to verify expected directions (higher scores associated with shorter calving intervals, lower replacement pressure, and more favorable herd structure), consistent with the environmental-efficiency interpretation.

\textbf{Predictive utility:} Finally, STGNN forecasting performance was compared when predicting $I_n^{\ast}$ versus alternative targets (individual pillar scores or unweighted raw-indicator averages). The composite index yields higher explanatory power and smoother trajectories, indicating that weighted pillar aggregation reduces noise and produces a more stable signal for spatio-temporal learning. Collectively, these checks show that the scoring framework is robust to reasonable perturbations, stable across counties and years, and consistent with sustainability-relevant operational proxies.

\section{STGNN for Sustainability Score Forecasting}

\begin{figure}[t!]
    \centering
    \includegraphics[width=1\linewidth]{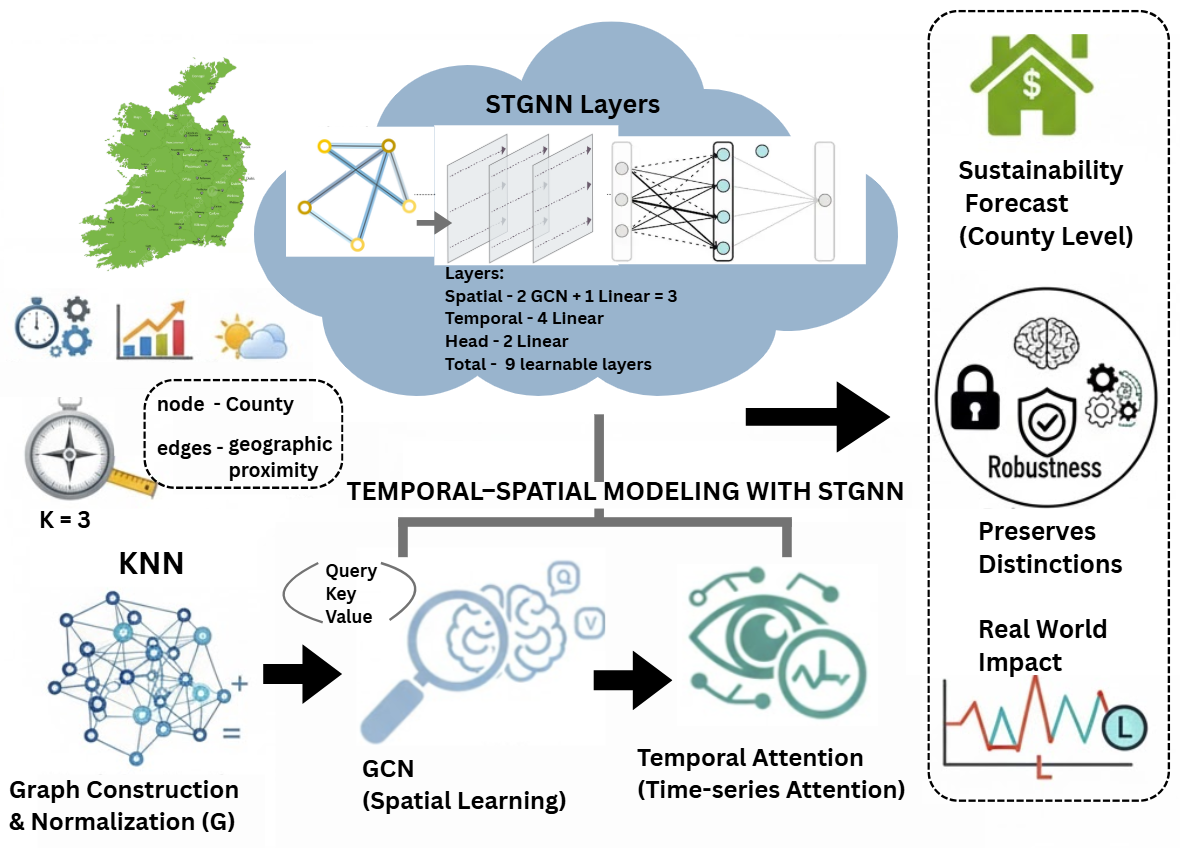}
    \caption{STGNN Architecture}
    \label{fig:placeholder}
\end{figure}

This section describes how the composite sustainability index is embedded into a spatio–temporal learning framework to generate county-level forecasts. Building on the PCA-derived pillars, the weighted sustainability scoring scheme, and the augmented dataset, the problem is formulated as regression on a sequence of graphs in which nodes represent counties, edges represent geographic proximity, and node features summarize both operational indicators and lagged sustainability scores. First, a spatial graph is constructed to encode geographic neighborhoods and to provide a stable substrate for graph-based message passing. Second, a STGNN is defined that combines graph convolutional layers for spatial aggregation with a temporal attention mechanism that learns how past years influence future sustainability trajectories. Finally, the resulting architecture is trained end-to-end to forecast the sustainability index, providing a scalable tool that links herd-level management patterns and environmental context to medium-term regional sustainability outcomes.

\subsection{Spatial Graph Construction}
\label{subsec:spatialgraph}
To model inter-county coupling,  a spatial graph was constructed in which counties are nodes and edges encode geographic proximity. Let $\mathcal{C}=\{1,\dots,N_c\}$ be the set of counties ($N_c=26$). Each county $i\in\mathcal{C}$ is associated with centroid coordinates $(\mathrm{lat}_i,\mathrm{lon}_i)$ in degrees. The geodesic distance between counties $i$ and $j$ is denoted $d_{ij}$ and computed via a great-circle metric (haversine), yielding a symmetric distance matrix $D=[d_{ij}]\in\mathbb{R}^{N_c\times N_c}$ with $d_{ii}=0$ and $d_{ij}=d_{ji}$.

A $k$-nearest-neighbor (kNN) graph was formed by connecting each county to its $k$ closest counties in geodesic distance. Specifically, the neighbor set of county $i$ is
$\mathcal{N}_k(i)=\{j\in\mathcal{C}\setminus\{i\}: d_{ij}\ \text{is among the $k$ smallest distances from $i$}\}$,
with $k=3$ in the experiments performed for this study. The undirected adjacency matrix $A\in\{0,1\}^{N_c\times N_c}$ is obtained by symmetrization:
$A_{ij}=1$ if $j\in\mathcal{N}_k(i)$ or $i\in\mathcal{N}_k(j)$, and $A_{ij}=0$ otherwise, ensuring $A_{ij}=A_{ji}$ and $A_{ii}=0$. The resulting graph is $\mathcal{G}=(\mathcal{V},\mathcal{E})$ with node set $\mathcal{V}=\mathcal{C}$ and edge set $\mathcal{E}=\{(i,j):A_{ij}=1\}$.

For implementation in message-passing GNN layers, edges were stored as $\mathrm{edge\_index}\in\mathbb{N}^{2\times E}$, where $E=|\mathcal{E}|$ and each undirected edge $(i,j)$ is represented by two directed pairs $(i,j)$ and $(j,i)$. Let $D_g\in\mathbb{R}^{N_c\times N_c}$ be the degree matrix with $(D_g)_{ii}=\sum_{j=1}^{N_c}A_{ij}$. Self-loops were addded via $\hat{A}=A+I$ (with $I$ the $N_c\times N_c$ identity) and define the corresponding degree matrix $\hat{D}$ analogously. The symmetrically normalized adjacency used for stable graph convolution is
\begin{equation}
\tilde{A}=\hat{D}^{-\frac{1}{2}}\hat{A}\hat{D}^{-\frac{1}{2}},
\end{equation}
which controls numerical scale during message passing and improves optimization stability in the STGNN.

\subsubsection{Graph Connectivity, Confidence, and Robustness}
 The kNN  was examined to ensure that the spatial graph is both meaningful and numerically stable. Let $\kappa(k)$ denote the number of connected components induced by the kNN parameter $k$. Small $k$ can fragment the graph (large $\kappa$), whereas large $k$ yields an overly dense graph that risks spatial over-smoothing. With $k=3$, the symmetrized kNN construction guarantees bounded sparsity: each node degree satisfies $1\le \deg(i)\le 2k$ for all $i\in\mathcal{C}$, implying no isolated counties ($\deg(i)=0$) while keeping message passing local and computationally efficient. Neighbor robustness to small coordinate perturbations was also assessed. For county $i$, let $d_i^{(1)}\le\dots\le d_i^{(N_c-1)}$ be the sorted distances to other counties. The margin
$\Delta_i=d_i^{(k+1)}-d_i^{(k)}$
separates the farthest included neighbor from the closest excluded neighbor. If coordinate uncertainty is small relative to $\Delta_i$, the neighbor set $\mathcal{N}_k(i)$ is unchanged. In practice, centroid geocoding errors are negligible compared to inter-county separations, so the resulting adjacency is stable under reasonable perturbations.

Finally, numerical stability  of graph convolutions was verified through the normalized adjacency $\tilde{A}$ (Section~\ref{subsec:spatialgraph}). For a single time step, let $H^{(0)}\in\mathbb{R}^{N_c\times F}$ be the node-feature matrix with $F$ features per county. A typical GCN update is
$H^{(\ell+1)}=\sigma\!\big(\tilde{A}H^{(\ell)}W^{(\ell)}\big)$,
where $\ell$ is the layer index, $W^{(\ell)}$ is a learnable weight matrix, and $\sigma(\cdot)$ is a nonlinearity. Because $\tilde{A}$ is symmetrically normalized, its spectrum is bounded, which prevents uncontrolled amplification under repeated message passing and promotes stable, locality-preserving smoothing. Combined with the geographic grounding of edges (geodesic proximity), bounded degrees, and margin-based robustness, these properties provide confidence that the constructed graph is an appropriate foundation for subsequent spatio-temporal learning.

\subsection{Temporal-Spatial Modeling with STGNN}
\label{subsec:stgnn}
To jointly capture temporal persistence and spatial coupling in county sustainability,  a spatio-temporal graph neural network (STGNN) operating on a sequence of county graphs was employed. Nodes correspond to counties $i\in\mathcal{C}=\{1,\dots,N_c\}$ and edges follow the proximity graph in Section~\ref{subsec:spatialgraph}. Let $t\in\mathcal{T}=\{1,\dots,T\}$ index training years and let $y_{i,t}$ denote the rescaled sustainability score (cf. \eqref{eq:scaled_index}) for county $i$ at year $t$.

To encode temporal persistence directly in the node attributes, A standardized one-year lag of the sustainability score was included as an additional feature. Specifically, the lag feature was defined
\begin{equation}
x^{\mathrm{lag}}_{i,t}=\frac{y_{i,t-1}-\bar{y}}{\sigma_y},
\label{eq:lag_feature}
\end{equation}
where $\bar{y}$ and $\sigma_y$ are the sample mean and standard deviation of $\{y_{i,t-1}\}$ over all counties and training years. This normalization places the lag on the same scale as the other standardized indicators, preventing it from dominating the learning dynamics.

Let $F_0$ be the number of standardized operational/pillar features available at each $(i,t)$, collected in $\mathbf{u}_{i,t}\in\mathbb{R}^{F_0}$. The augmented node feature vector is then $\mathbf{x}_{i,t}=[\mathbf{u}_{i,t}^{\top}\;\;x^{\mathrm{lag}}_{i,t}]^{\top}\in\mathbb{R}^{F}$ with $F=F_0+1$. Stacking all counties at time $t$ yields the node-feature matrix
$X_t=[\mathbf{x}_{1,t}^{\top},\dots,\mathbf{x}_{N_c,t}^{\top}]^{\top}\in\mathbb{R}^{N_c\times F}$,
which serves as the STGNN input for year $t$.

\subsubsection{Spatial Graph Convolution}
Spatial interactions are learned via graph convolutions on the county graph $\mathcal{G}=(\mathcal{V},\mathcal{E})$ using the symmetrically normalized adjacency $\tilde{A}\in\mathbb{R}^{N_c\times N_c}$ (Section~\ref{subsec:spatialgraph}). At year $t$, the initial node-feature matrix is $H_t^{(0)}=X_t\in\mathbb{R}^{N_c\times F}$. A GCN layer produces spatial embeddings by neighborhood aggregation:
\begin{equation}
H_t^{(\ell+1)}=\sigma\!\big(\tilde{A}H_t^{(\ell)}W^{(\ell)}\big), \qquad \ell=0,\dots,L_{\mathrm{sp}}-1,
\label{eq:gcn_layer}
\end{equation}
where $W^{(\ell)}$ is a learnable weight matrix, $\sigma(\cdot)$ is an element-wise nonlinearity (ReLU), $L_{\mathrm{sp}}$ is the number of spatial layers, and $H_t^{(\ell)}\in\mathbb{R}^{N_c\times F^{(\ell)}}$ with $F^{(0)}=F$. The output of the final spatial block is $H_t=H_t^{(L_{\mathrm{sp}})}\in\mathbb{R}^{N_c\times F_{\mathrm{sp}}}$, and the embedding of county $i$ at time $t$ is the row vector $\mathbf{h}_{i,t}=H_t(i,:)\in\mathbb{R}^{F_{\mathrm{sp}}}$. Because $\tilde{A}$ is normalized, repeated multiplication acts as controlled smoothing over local neighborhoods, improving numerical stability and limiting feature scale drift as depth increases.

\subsubsection{Temporal Attention Mechanism}
Temporal dependence is modeled by attention over the sequence of spatial embeddings for each county. For county $i$, the embedding history is $\{\mathbf{h}_{i,\tau}\}_{\tau=1}^{T}$ with $\mathbf{h}_{i,\tau}\in\mathbb{R}^{F_{\mathrm{sp}}}$. Embeddings were projected to query, key, and value spaces:
$\mathbf{q}_{i,t}=\mathbf{h}_{i,t}W_Q$, $\mathbf{k}_{i,\tau}=\mathbf{h}_{i,\tau}W_K$, and $\mathbf{v}_{i,\tau}=\mathbf{h}_{i,\tau}W_V$,
where $W_Q,W_K\in\mathbb{R}^{F_{\mathrm{sp}}\times d}$, $W_V\in\mathbb{R}^{F_{\mathrm{sp}}\times d_v}$ are learnable matrices, and $d$ and $d_v$ are the query/key and value dimensions. The attention weight assigned to historical year $\tau$ when forming the representation at time $t$ is
\begin{equation}
\alpha_{i,t,\tau}=\frac{\exp\!\left(\mathbf{q}_{i,t}^{\top}\mathbf{k}_{i,\tau}/\sqrt{d}\right)}
{\sum_{\ell=1}^{T}\exp\!\left(\mathbf{q}_{i,t}^{\top}\mathbf{k}_{i,\ell}/\sqrt{d}\right)},
\label{eq:attention_weights}
\end{equation}
so that $\sum_{\tau=1}^{T}\alpha_{i,t,\tau}=1$. The temporally attended embedding is then
\begin{equation}
\mathbf{h}^{\mathrm{temp}}_{i,t}=\sum_{\tau=1}^{T}\alpha_{i,t,\tau}\mathbf{v}_{i,\tau}\in\mathbb{R}^{d_v}.
\label{eq:temp_embedding}
\end{equation}
The scaling by $1/\sqrt{d}$ stabilizes the softmax as $d$ increases. In practice, attention was restricted to a recent window of given length $L_{\mathrm{temp}}\le T$ by computing \eqref{eq:attention_weights}-\eqref{eq:temp_embedding} over $\tau\in\{t-L_{\mathrm{temp}}+1,\dots,t\}$ to emphasize recent dynamics and reduce computation.

\subsubsection{Regression Head and Training Objective}

The temporally attended representation $\mathbf{h}_{i,t}^{\text{temp}}$ is fed into a regression head that maps the spatio-temporal embedding to a predicted sustainability score. The regression head is implemented as a small feed-forward network with one or more fully connected layers and nonlinear activations. Denoting the regression mapping by $f_{\text{head}}(\cdot)$, the predicted score for county $i$ at time $t$ is
\begin{equation}
\hat{y}_{i,t} = f_{\text{head}}(\mathbf{h}_{i,t}^{\text{temp}}),
\label{eq:stgnn_pred}
\end{equation}
where $\hat{y}_{i,t}$ is a scalar approximating the true rescaled Sustainability Score $y_{i,t}$.

Training is performed by minimizing the mean squared error (MSE) between predicted and true scores across all counties and training time steps, augmented with weight regularization to promote generalization. Let $\mathcal{T}_{\text{train}} \subseteq \mathcal{T}$ denote the set of years used for training and let $\Theta$ denote the collection of all trainable parameters in the GCN layers, temporal attention mechanism, and regression head. The loss function is
\begin{equation}
\mathcal{L}_{\text{STGNN}}(\Theta) = 
\frac{1}{|\mathcal{T}_{\text{train}}| N_c}
\sum_{t \in \mathcal{T}_{\text{train}}} \sum_{i=1}^{N_c}
\left( y_{i,t} - \hat{y}_{i,t} \right)^{2}
+ \lambda \, \|\Theta\|_{2}^{2},
\label{eq:stgnn_loss}
\end{equation}
where $\lambda \geq 0$ is a regularization coefficient and $\|\Theta\|_{2}^{2}$ denotes the sum of squared parameters ($l_2$ regularization). The loss \eqref{eq:stgnn_loss} is minimized using stochastic gradient-based optimization (for example, Adam) with mini-batches defined over the temporal dimension or over county-time pairs.

\subsubsection{Convergence, Confidence, and Robustness}
The STGNN is fully differentiable and is trained by minimizing a nonnegative objective (cf. \eqref{eq:stgnn_loss}), so optimization proceeds by iteratively reducing the training loss to a stationary point under standard learning-rate and optimizer conditions. Numerical stability is promoted by two normalization mechanisms: (i) the spatial block uses the normalized adjacency $\tilde{A}$, which keeps message-passing magnitudes controlled, and (ii) the temporal block uses softmax attention weights $\alpha_{i,t,\tau}$ in \eqref{eq:attention_weights}, which form a convex weighting over history and therefore cannot arbitrarily amplify intermediate representations.

Generalization was improved via regularization during training. Dropout applied in the GCN and regression head reduces co-adaptation by stochastically masking hidden units, while $\ell_2$ weight decay in \eqref{eq:stgnn_loss} discourages overly complex parameterizations. Temporal robustness is enhanced by combining the standardized lag feature $x^{\mathrm{lag}}_{i,t}$ in \eqref{eq:lag_feature} with attention: if recent dynamics dominate, attention concentrates on $\tau\approx t$; if longer-term trends matter, it distributes mass across a broader range without requiring a manually fixed memory length. Spatial robustness follows from the sparse kNN graph and bounded smoothing of the GCN stack: with small spatial depth $L_{\mathrm{sp}}$, the model captures regional coherence while avoiding over-smoothing (collapse of node embeddings to near-constants) and preserving county-level differentiation.

\subsection{Deployment as a Decision Support Tool}
After training, inference is lightweight: for each year, $L_{\mathrm{sp}}$ graph-convolution layers were applied followed by attention over a short history window, making the approach suitable for routine updates in analytics dashboards. Because inputs are standardized operational indicators that are routinely collected through digital herd recording (e.g., EID-linked events, breeding and health records, and centralized databases), new county-year observations can be ingested, normalized, and propagated through the fixed STGNN to refresh forecasts and uncertainty summaries. The combination of spatial aggregation and adaptive temporal weighting yields forecasts that are stable to short-term noise yet responsive to sustained management changes. The framework is also extensible: additional node- or edge-level streams (e.g., local weather, soil proxies, or direct emissions sensor measurements) can be appended to the feature vectors to move from proxy-based sustainability toward explicit environmental attribution. This fusion would enable scenario-aware decision support that jointly reflects management actions and measured environmental conditions, improving the operational relevance of forecasts for farm advisors and policy makers.

\section{Numerical Results Formulation}

This section formalizes how the numerical results reported in Section~V were generated, namely (i) county-level sustainability forecasts for 2026-2030, (ii) uncertainty bands via Monte Carlo simulation, and (iii) policy-relevant counterfactual trajectories under targeted interventions. Starting from standardized county-year feature vectors, inputs were extrapolated beyond the observed window, temporal history was compressed using an attention mechanism, and forecasts were produced with the trained STGNN. Forecast risk was then quantified by perturbing predictions to obtain empirical prediction intervals, and intervention impacts were evaluated by modifying a small set of high-leverage features for selected counties while keeping the remaining counties on baseline trajectories to preserve network context.

\subsection{Forecasting Future Sustainability (2026-2030)}
To generate county-level forecasts, a rolling temporal input was constructed for each county using the historical window 2021-2025. Let $i\in\{1,\dots,N\}$ index counties ($N=26$) and $t$ denote the year. For county $i$ at year $t$, the $m$-dimensional feature vector was defined of standardized operational indicators as
\begin{equation}
X_{i,t} = [x^{(1)}_{i,t}, x^{(2)}_{i,t}, \dots, x^{(m)}_{i,t}], \quad t\in\{2021,\dots,2025\},
\end{equation}
where $x^{(j)}_{i,t}$ is the value of feature $j\in\{1,\dots,m\}$ for county $i$ at year $t$. An input sequence was formed of length $n$ years (here $n=5$) as $X_i=[X_{i,t-n+1},\dots,X_{i,t}]\in\mathbb{R}^{n\times m}$.

To extend features beyond the observation horizon, each feature was extrapolated independently using a per-county linear trend fitted to the historical values, and Gaussian noise was injected to represent variability:
\begin{equation}
\hat{x}^{(j)}_{i,t+1} = a^{(j)}_{i}\, t + b^{(j)}_{i} + \epsilon,\qquad \epsilon\sim\mathcal{N}(0,\sigma^2),
\end{equation}
where $\hat{x}^{(j)}_{i,t+1}$ is the projected value of feature $j$ for county $i$ at year $t+1$, $a^{(j)}_{i}$ and $b^{(j)}_{i}$ are the fitted slope and intercept for that county-feature pair, $\epsilon$ is zero-mean Gaussian noise, and $\sigma^2$ is the noise variance. The projected features are subsequently clipped to a predefined feasible range (e.g., $[0,1]$ after normalization) to avoid physically implausible values.

The projected sequence is then processed by a temporal attention block that learns which years are most informative for prediction. Given $X_i\in\mathbb{R}^{n\times m}$, computations were performed
\begin{equation}
Q_i = X_iW_Q,\quad K_i = X_iW_K,\quad V_i = X_iW_V,
\end{equation}
where $Q_i,K_i,V_i\in\mathbb{R}^{n\times d}$ are the query, key, and value matrices, $d$ is the attention embedding dimension, and $W_Q,W_K,W_V\in\mathbb{R}^{m\times d}$ are learnable projection matrices. Attention weights for predicting at time index $t$ (within the $n$-year window) are computed as
\begin{equation}
\alpha_{i,t,k}=\frac{\exp(Q_{i,t}K_{i,k}^{\top})}{\sum_{l=1}^{n}\exp(Q_{i,t}K_{i,l}^{\top})},
\end{equation}
where $Q_{i,t}\in\mathbb{R}^{1\times d}$ is the query vector at time index $t$, $K_{i,k}\in\mathbb{R}^{1\times d}$ is the key vector at time index $k$, and $\alpha_{i,t,k}$ is the normalized attention weight. The temporally attended representation is
\begin{equation}
\tilde{X}_{i,t}=\sum_{k=1}^{n}\alpha_{i,t,k}V_{i,k},
\end{equation}
where $V_{i,k}\in\mathbb{R}^{1\times d}$ is the value vector at time index $k$ and $\tilde{X}_{i,t}\in\mathbb{R}^{1\times d}$ aggregates the most predictive historical information.

Finally, the attended representation is mapped to the sustainability forecast via the trained STGNN:
\begin{equation}
\hat{y}_{i,t}=f_{\mathrm{STGNN}}(\tilde{X}_{i,t}),
\end{equation}
where $\hat{y}_{i,t}$ is the predicted sustainability score for county $i$ at year $t$ and $f_{\mathrm{STGNN}}(\cdot)$ denotes the learned STGNN predictor (including spatial graph operations and the output regression head). This procedure was applied auto-regressively for $t=2026,\dots,2030$, with each predicted $\hat{y}_{i,t}$ appended to the county history to condition subsequent forecasts.

\subsection{Monte Carlo-Based Uncertainty Estimation}
Forecast uncertainty was quantified for the 2026-2030 horizon using a Monte Carlo procedure applied to representative counties (Cork and Dublin). Let $i$ index the county and $t\in\{2026,\dots,2030\}$ denote the forecast year. The deterministic STGNN baseline prediction is denoted by $\hat{y}^{\mathrm{base}}_{i,t}$, where $\hat{y}$ is the predicted sustainability score under unchanged (baseline) feature conditions. To model small, irreducible variability around the baseline, $S$ perturbed realizations were generated by additive Gaussian noise, $\epsilon^{(s)}_{i,t}\sim\mathcal{N}(0,\sigma^2)$, where $s\in\{1,\dots,S\}$ is the Monte Carlo index, $S=100$ is the number of trials, and $\sigma$ is the noise standard deviation (set to $0.001\times 10$ in the experiments). Each simulated trajectory is
\begin{equation}
y^{(s)}_{i,t}=\hat{y}^{\mathrm{base}}_{i,t}+\epsilon^{(s)}_{i,t}.
\end{equation}

The Monte Carlo mean provides the expected forecast,
\begin{equation}
\bar{y}_{i,t}=\frac{1}{S}\sum_{s=1}^{S}y^{(s)}_{i,t},
\end{equation}
and uncertainty is summarized by an empirical interval computed from the simulated distribution at each year $t$. Specifically, the 5th and 95th percentiles were reported, denoted $q_{0.05}(y_{i,t})$ and $q_{0.95}(y_{i,t})$, which form a 90\% prediction band around $\bar{y}_{i,t}$. In Figs.~\ref{fig:montecarloCork}-\ref{fig:montecarlodublin}, $\bar{y}_{i,t}$ was visualized as a line and the corresponding percentile band was visualized as a shaded region, providing a compact, decision-relevant view of the forecast trend and its plausible bounds under baseline conditions.

\subsection{Counterfactual Scenario Simulations}
To evaluate the effect of targeted management interventions, counterfactual simulations were performed using the trained STGNN while preserving the spatial graph structure. Let $i\in{1,\dots,N}$ index counties and $t\in{2026,\dots,2030}$ denote forecast years. For each county $i$, the baseline (business-as-usual) projected feature vector is $\hat{X}_{i,t}=[\hat{x}^{(1)}_{i,t},\dots,\hat{x}^{(m)}_{i,t}]$, where $\hat{x}^{(j)}_{i,t}$ is the extrapolated value of feature $j$ under baseline conditions.

 Interventions were applied only to selected counties (Monaghan and Kerry) and to a small subset of high-impact features, while all other counties follow their baseline trajectories. For an intervened county $i$, the counterfactual feature value for feature $j$ is
\begin{equation}
\tilde{x}^{(j)}_{i,t}=\hat{x}^{(j)}_{i,t}\,(1+\delta_{i,j}),
\end{equation}
where $\tilde{x}^{(j)}_{i,t}$ is the modified feature value and $\delta_{i,j}$ is the intervention strength (positive for an increase and negative for a decrease). In the experiments, Monaghan uses $\delta$ values corresponding to: Recycled Cows (\%) $+0.15$, Calving Interval (days) $-0.05$, and Cows Culled in Period (\%) $-0.05$; Kerry uses: Recycled Cows (\%) $+0.08$, Calving Interval (days) $-0.05$, and Cows Culled in Period (\%) $-0.02$. The resulting counterfactual feature vector is $\tilde{X}_{i,t}$.

For each forecast year, baseline and counterfactual sustainability scores were computed via the same trained predictor $f_{\mathrm{STGNN}}(\cdot)$:
\begin{equation}
\hat{y}^{\mathrm{base}}_{i,t}=f_{\mathrm{STGNN}}(\hat{X}_{i,t}), \qquad
\tilde{y}^{\mathrm{scen}}_{i,t}=f_{\mathrm{STGNN}}(\tilde{X}_{i,t}),
\end{equation}
where $\hat{y}^{\mathrm{base}}_{i,t}$ and $\tilde{y}^{\mathrm{scen}}_{i,t}$ denote the baseline and scenario sustainability scores, respectively. The intervention effect is summarized by the year-wise uplift $\Delta y_{i,t}=\tilde{y}^{\mathrm{scen}}_{i,t}-\hat{y}^{\mathrm{base}}_{i,t}$. Simulation was executed auto-regressively  from 2026 to 2030, updating each county’s input sequence with the newly predicted score so that the intervention impact can accumulate over time. This design yields a policy-relevant ``what-if'' analysis: by isolating changes to specific management levers in selected counties while holding the broader network evolution fixed, $\Delta y_{i,t}$ quantifies the expected medium-term sustainability gains attributable to targeted improvements in reproductive performance, culling efficiency, and welfare-related outcomes.

\section{Numerical Results \& Discussions}

This section evaluates the proposed sustainability assessment and forecasting pipeline, with emphasis on STGNN for county-level modeling. The STGNN captures both temporal persistence and spatial spillovers across neighboring counties, yielding stable forecasts for 2026-2030 and uncertainty-aware prediction bands via Monte Carlo simulations. The counterfactual experiments further demonstrate how targeted improvements in key operational drivers translate into measurable gains in the composite sustainability score, providing a direct, policy-relevant interpretation. Comparisons against RNN, LSTM, FFNN, and GKR confirm that the STGNN produces smoother and more physically plausible trajectories, avoiding abrupt oscillations that are difficult to justify in real dairy systems where management and biological responses evolve gradually.

\subsection{Generation and Evaluation of Synthetic Data Using VAE}
\begin{figure}[t!]
    \centering
    \includegraphics[width=0.7\columnwidth]{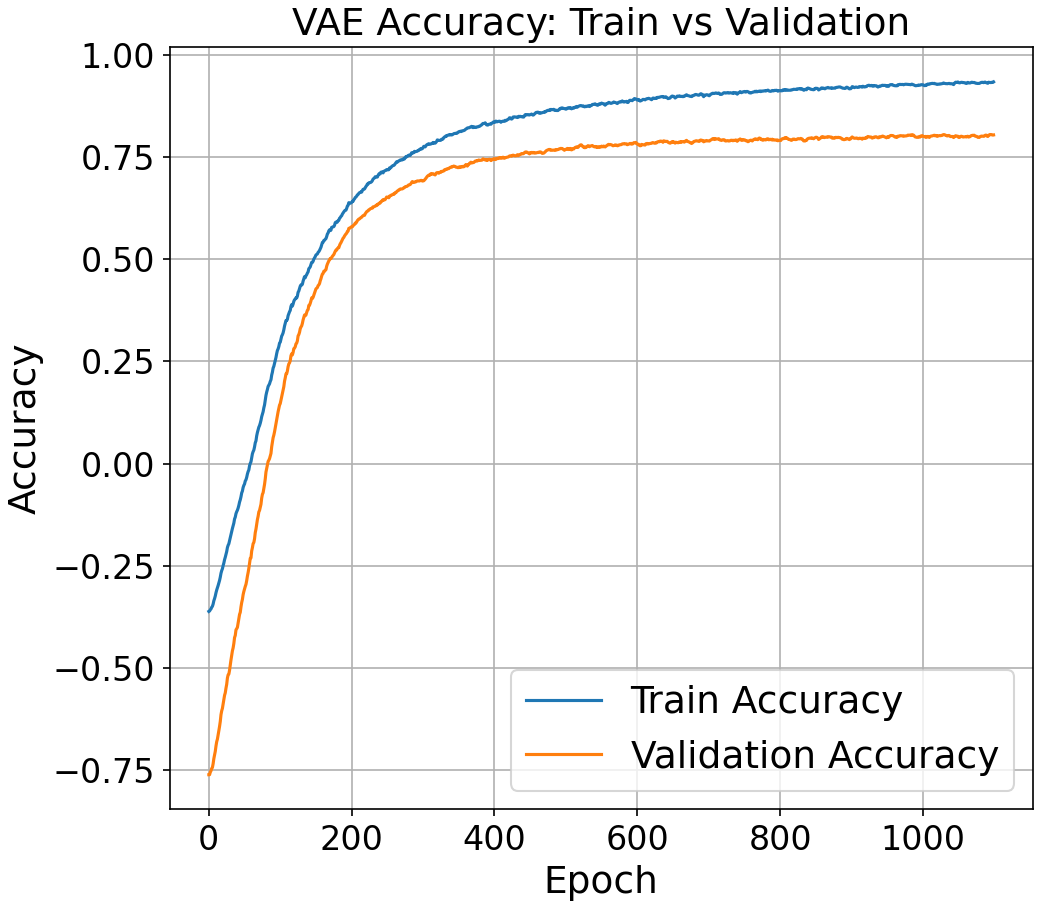}
    \caption{Training and Validation Accuracy of VAE}
    \vspace{-5mm}
    \label{fig:vae_acc}
\end{figure}

A variational autoencoder (VAE) was used to augment the limited farm-level dataset while preserving its multivariate structure. Fig.~\ref{fig:vae_acc} reports the reconstruction accuracy over training, showing convergence to $\approx 0.95$ on the training set and $\approx 0.81$ on the validation set. The high training accuracy indicates that the latent representation captures the dominant correlations among herd-level indicators, while the lower (yet strong) validation accuracy reflects realistic generalization limits given the modest sample size and inherent variability across farms and years. From a physical standpoint, this behavior is expected: operational metrics such as fertility, health, and herd structure exhibit consistent co-movements, but also include farm-specific noise and management heterogeneity that cannot be perfectly reconstructed from compressed latent factors. The VAE-generated samples substantially expanded the dataset from 130 original rows to 650 total rows by adding 520 synthetic rows (a $4\times$ increase). This augmentation improves temporal coverage and reduces sparsity without changing the marginal feature definitions, providing a more reliable basis for downstream learning. In practice, the enlarged dataset stabilizes model training, supports more robust uncertainty quantification, and enables scenario exploration while retaining consistency with observed herd-level patterns.

\subsection{Farm-Level Pillars and Computation of Composite Sustainability Scores}
\begin{figure}[h!]
\centering
\includegraphics[width=1\columnwidth]{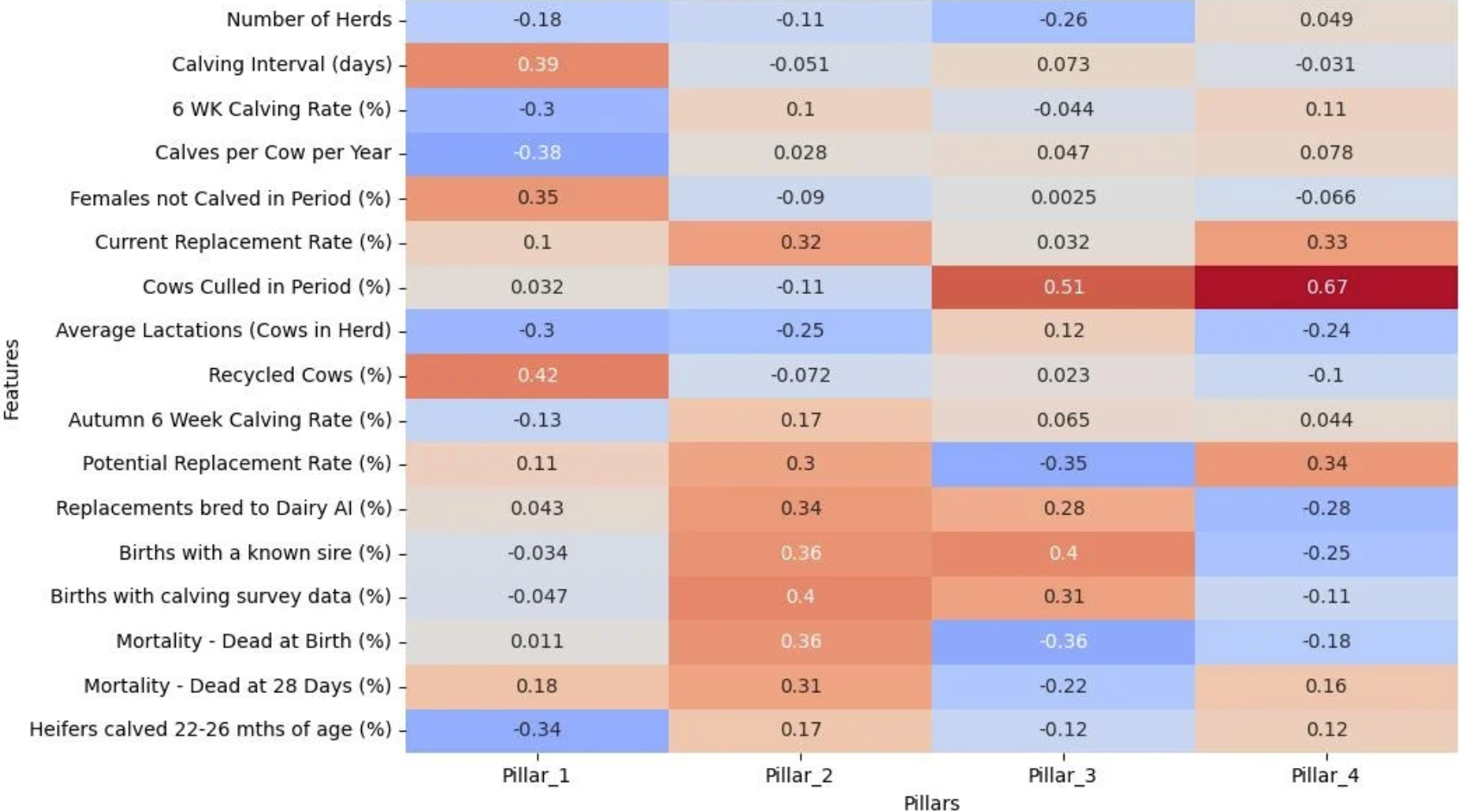}
\vspace{-5mm}
\caption{PCA feature loadings illustrating the relative contribution of each variable to the four pillars.}

\label{fig:pca}
\end{figure}

To obtain an interpretable and compact representation of dairy-farm operations, the herd-level indicators were consolidated into four farm-level pillars: \emph{Reproductive Efficiency}, \emph{Genetic Management}, \emph{Herd Health}, and \emph{Herd Management}. These pillars form the basis of the composite sustainability score, intended to summarize operational performance in a manner that is informative for environmental outcomes (e.g., pathways influencing methane emissions and overall carbon footprint) \cite{Clasen2024}. The pillar construction is data-driven: PCA was applied to the full feature set, and the resulting loading matrix is visualized in Fig.~\ref{fig:pca}. The heatmap highlights which variables dominate each component, enabling a physically meaningful grouping. For instance, fertility-related variables (e.g., calving interval and calving rates) load strongly on the reproductive component, reflecting the biological reality that tighter calving patterns reduce unproductive days and replacement pressure. Similarly, AI usage, sire recording, and replacement-related variables concentrate in the genetics/management components, consistent with how breeding decisions determine herd productivity and persistence. Health- and loss-related metrics (e.g., culling and mortality) emerge as key drivers of the health pillar, capturing welfare and avoidable inefficiencies that also propagate to environmental burdens through wasted inputs and increased replacement demand.

Following pillar identification, the highest-loading variables for each pillar (Table~\ref{tab:topfeatures}) were selected and normalized to a common $[0,1]$ scale to ensure comparability across farms and regions. The normalized pillar scores were then aggregated into a single composite sustainability score using variance-based weights derived from PCA (Table~\ref{tab:variance}). These weights quantify the information retained by each pillar and provide a principled aggregation that emphasizes components explaining more variability in herd performance. As shown in Table~\ref{tab:variance}, the four pillars together account for essentially all meaningful variance (weights sum to $\approx 1$), indicating that the resulting composite score is a faithful low-dimensional surrogate of the original feature space and is therefore suitable for downstream forecasting and counterfactual analysis. The variables with the largest absolute PCA loadings (i.e., dominant contributors) for each pillar are summarized in Table~\ref{tab:topfeatures}. The variance contributions (aggregation weights) of the four PCA-derived pillars are listed in Table~\ref{tab:variance}. The near-unit sum indicates that the pillars retain essentially all informative variability required for composite scoring and subsequent modeling.

\begin{table}[t!]
\centering
\begin{tabular}{c p{11cm}}
\hline
\textbf{Pillar} & \textbf{Top Contributing Features} \\
\hline
Pillar 1 & Recycled Cows (\%), Calving Interval (days), Calves per Cow per Year, Females not Calved in Period (\%), Heifers calved 22-26 mths of age (\%) \\
\hline
Pillar 2 & Births with calving survey data (\%), Replacements bred to Dairy AI (\%), Births with a known sire (\%), Current Replacement Rate (\%), Potential Replacement Rate (\%) \\
\hline
Pillar 3 & Cows Culled in Period (\%), Mortality - Dead at Birth (\%), Potential Replacement Rate (\%), Births with a known sire (\%), Mortality - Dead at 28 Days (\%) \\
\hline
Pillar 4 & Cows Culled in Period (\%), Potential Replacement Rate (\%), Current Replacement Rate (\%), Replacements bred to Dairy AI (\%), Births with a known sire (\%) \\
\hline
\end{tabular}
\caption{Top contributing features for each PCA-derived pillar.}
\label{tab:topfeatures}
\end{table}

\begin{table}[t!]
\centering
\begin{tabular}{c p{5cm}}
\hline
\textbf{Pillar} & \textbf{Variance Contribution (Weight)} \\
\hline
Pillar 1 & 0.3205 ($\sim$32\%) \\
\hline
Pillar 2 & 0.2564 ($\sim$26\%) \\
\hline
Pillar 3 & 0.2308 ($\sim$23\%) \\
\hline
Pillar 4 & 0.1923 ($\sim$19\%) \\
\hline
\end{tabular}
\caption{Variance contribution of each PCA-derived pillar.}
\label{tab:variance}
\end{table}\vspace{-3mm}

\subsection{STGNN Model Training and Evaluation}
The STGNN was trained to forecast county-level sustainability scores using historical observations from 2021-2025 across 26 counties. Training converged in 44.3~s and achieved a best validation accuracy of 0.9186. Table~I summarizes the final regression performance: the STGNN attains $R^2=0.9866$ on training data and generalizes well with $R^2=0.9124$ on validation and $R^2=0.9072$ on the held-out 2025 test set, with low absolute errors (MAE $\approx 0.047$). These results indicate that the learned dynamics remain predictive under year-to-year variability.

\begin{figure}[t!]
    \centering
    \includegraphics[width=1\linewidth]{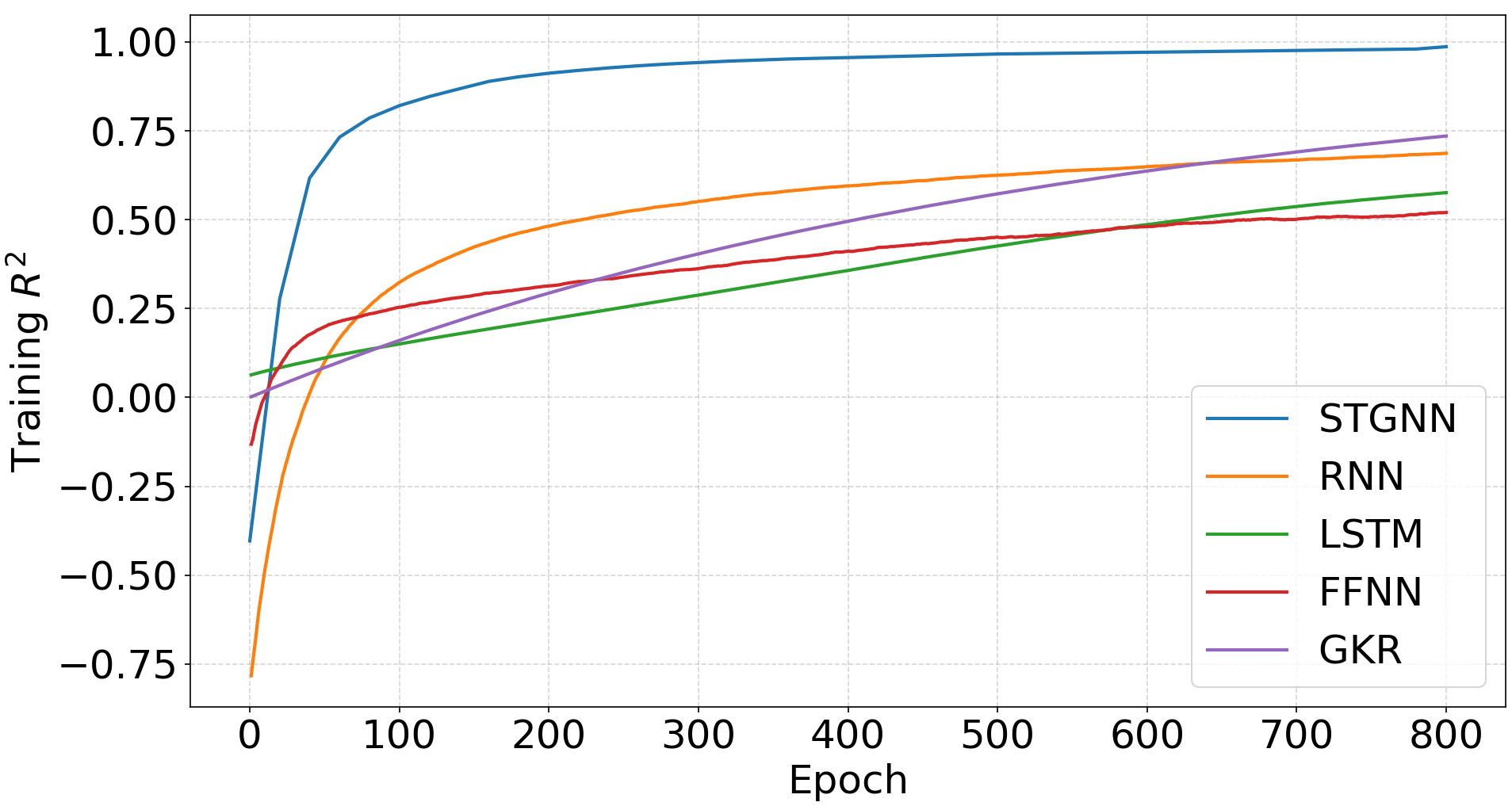}
    \caption{Training $R^2$ comparison for five models: STGNN, RNN, LSTM, FFNN, and GKR.}
    \label{fig:training accuracy}
    \vspace{-4mm}
\end{figure}

\begin{figure}[t!]
    \centering
    \includegraphics[width=1\linewidth]{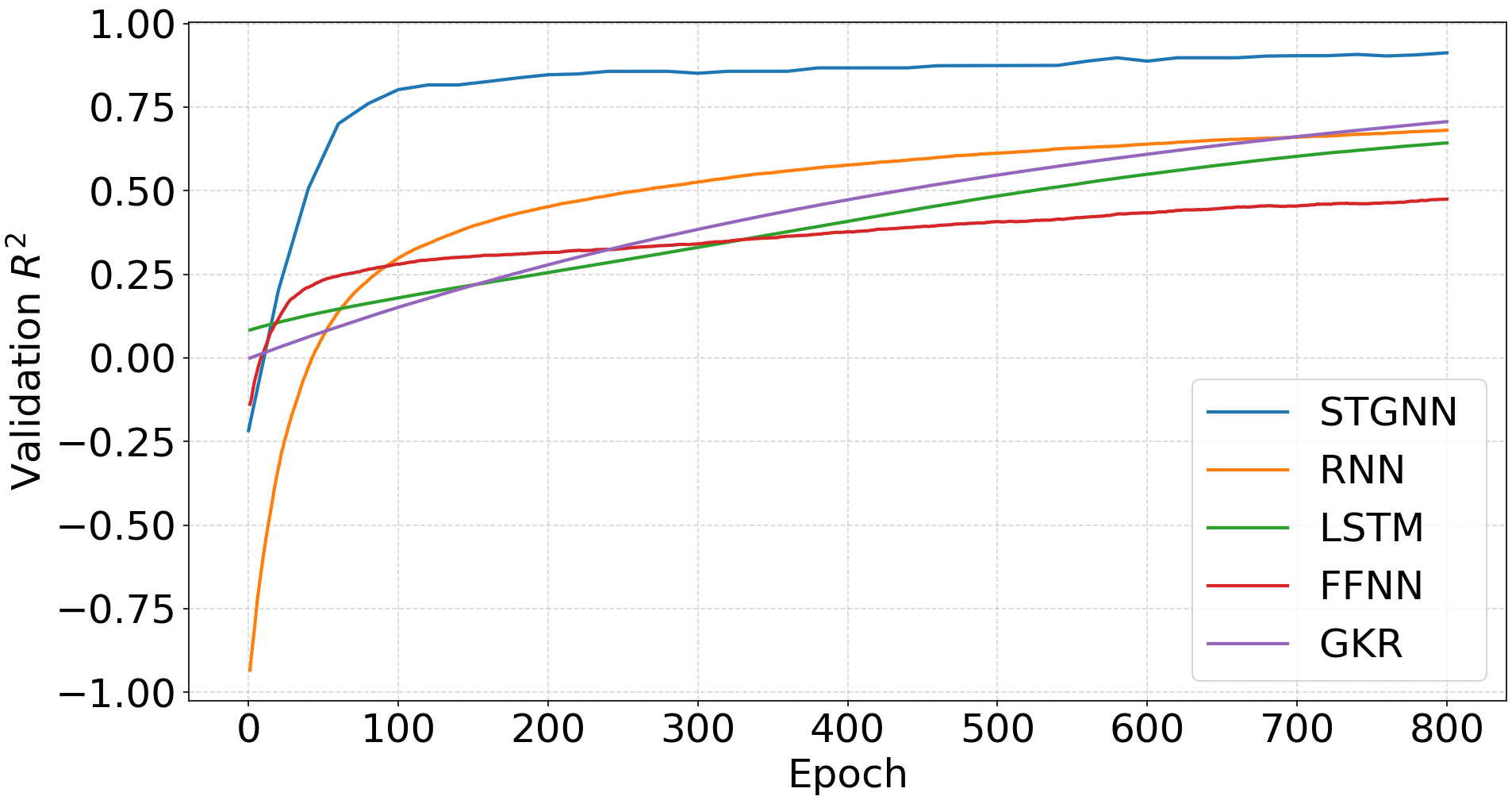}
    \caption{Validation $R^2$ comparison for five models: STGNN, RNN, LSTM, FFNN, and GKR.}
    \vspace{-5mm}
    \label{fig:validation accuracy}
\end{figure}

Figs.~\ref{fig:training accuracy}-\ref{fig:validation accuracy} compare learning curves across STGNN and four baselines (RNN, LSTM, FFNN, GKR) over 800 epochs using $R^2$ as the primary goodness-of-fit measure. In Fig.~\ref{fig:training accuracy}, the STGNN rapidly approaches a near-saturated training fit (around $0.98$), whereas the purely temporal models (RNN/LSTM) and the nonparametric regressor (GKR) plateau substantially lower, and the FFNN performs worst. This gap is consistent with the physical structure of the problem: sustainability at county scale is not driven solely by each county’s past trajectory, but also by spatially coupled effects (shared climate, infrastructure, advisory practices, and supply-chain conditions). By explicitly modeling county-to-county interactions through the graph component, the STGNN can explain variance that sequence-only (RNN/LSTM) or tabular (FFNN/GKR) models cannot represent.

Generalization is reflected in Fig.~\ref{fig:validation accuracy}. The STGNN maintains a smooth and stable validation curve around $R^2\approx 0.91$, with no abrupt degradation as training improves, indicating controlled overfitting and robust out-of-sample behavior. In contrast, the baselines stabilize at markedly lower validation performance (RNN $\sim 0.65$, LSTM $\sim 0.60$, GKR $\sim 0.73$, FFNN $\sim 0.49$), suggesting either limited capacity to represent spatio-temporal dependencies or insufficient inductive bias for gradual, coupled system dynamics. Overall, the consistent separation between the STGNN and alternatives across both training and validation supports its suitability for producing reliable, policy-relevant sustainability forecasts.

\begin{table}[t!]
\centering

\begin{tabular}{lccc}
\hline
\textbf{Dataset} & \textbf{$R^2$} & \textbf{MAE} & \textbf{MSE} \\
\hline
Training      & 0.9866 & 0.0095 & 0.000201 \\
Validation    & 0.9124 & 0.0473 & 0.003600 \\
Test (2025)   & 0.9072 & 0.0476 & 0.003788 \\
\hline
\end{tabular}
\caption{STGNN Performance Metrics}
\end{table}

\subsection{Sustainability Score Calculation and Historical Analysis (2021-2025)}

\begin{figure*}[t!]
    \centering
    \includegraphics[width=1.05\textwidth]{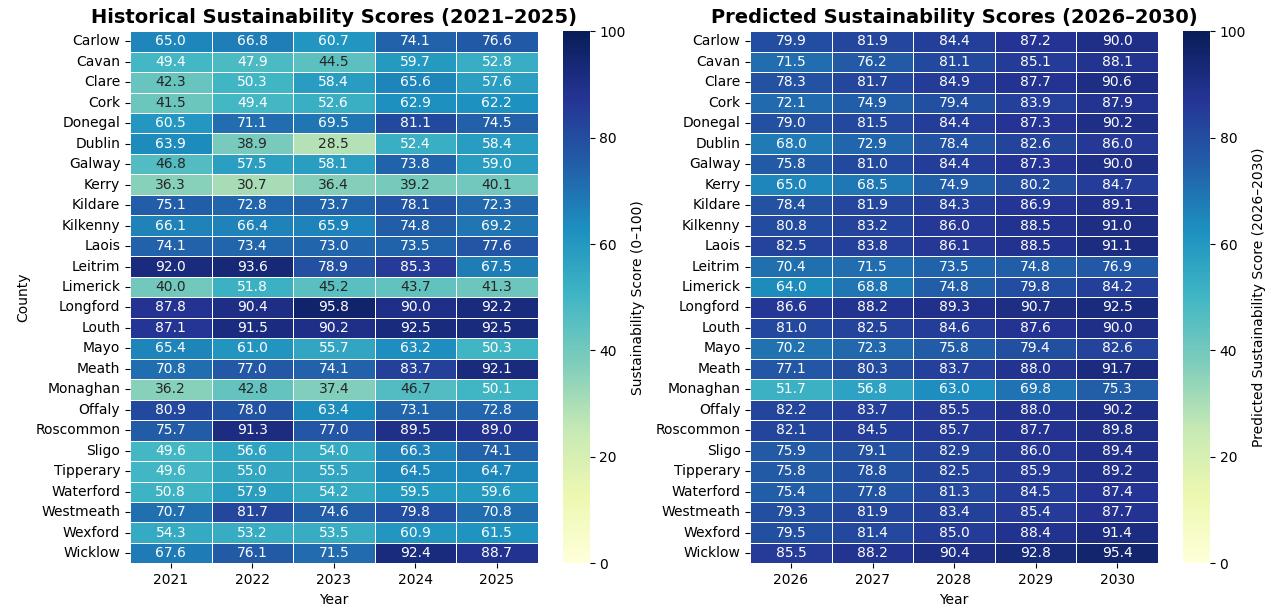}
    \caption{Historical (2021-2025) and predicted (2026-2030) dairy sustainability scores by STGNN.}
    \vspace{-5mm}
    \label{fig:sustainability score}
\end{figure*}

The Sustainability Score is built by first calculating four separate pillar scores, each representing a different dimension of dairy herd sustainability. For every pillar, a set of key features is selected based on their importance from PCA (Principal Component Analysis). Each feature within a pillar is multiplied by its PCA-derived weight, ensuring that more influential indicators contribute more strongly to the pillar score. The weighted values for the features in each pillar are then summed to produce Pillar 1, Pillar 2, Pillar 3, and Pillar 4 scores. Once all four pillars are computed, the overall Sustainability Score is calculated as the simple average of these four pillar scores, reflecting a balanced contribution from all sustainability dimensions. To make the score more interpretable and ensure comparability across counties and years, the raw Sustainability Score is then min-max scaled to a range of 10–100, where 10 represents the lowest sustainability performance observed in the dataset and 100 represents the highest. Finally, all values are rounded for clarity. This process produces a standardized, weighted sustainability metric that captures multi-dimensional herd performance in a single, interpretable score.
 The heatmap of sustainability scores for the period 2021 to 2025( Fig. \ref{fig:sustainability score}) presents the observed data used to train and evaluate the Spatio-Temporal Graph Neural Network (STGNN). Across this five-year span, nearly all Irish counties demonstrated consistent improvements in sustainability performance. For example, Carlow showed a steady rise from 65.0 in 2021 to 76.6 in 2025, while Dublin improved from 60.5 to 71.1. This upward trend indicates a broad and sustained enhancement in dairy sustainability across the country.

The spatial distribution of the scores highlights distinct regional disparities. High-performing counties such as Louth (92.5 in 2025) and Wicklow (88.7) consistently maintained leadership positions, reflecting relatively mature sustainability practices and more efficient production systems. In contrast, Kerry (40.1) and Limerick (41.3) recorded the lowest performance values, suggesting structural or resource-based limitations that may require targeted interventions. These observed patterns provide the empirical foundation for the forecasting component of the model, supplying both the temporal and spatial structure from which future trajectories are extrapolated.

\subsection{Predicted Sustainability Trajectory (2026-2030)}

Fig.~\ref{fig:sustainability score} reports the county-level sustainability heatmap for the historical period (2021-2025) and the STGNN forecast (2026-2030). The historical window shows noticeable inter-annual variability in several counties, consistent with short-term fluctuations in the underlying operational drivers (e.g., fertility performance, replacement pressure, and health-related losses) and the delayed response of management interventions. In contrast, the forecast exhibits smoother year-to-year evolution with a broadly increasing trajectory, indicating that the STGNN extracts the persistent trend component while attenuating transient perturbations.

\begin{figure}[t]
    \centering
\centering
    \begin{minipage}{0.45\textwidth}
        \centering
        \includegraphics[width=\linewidth]{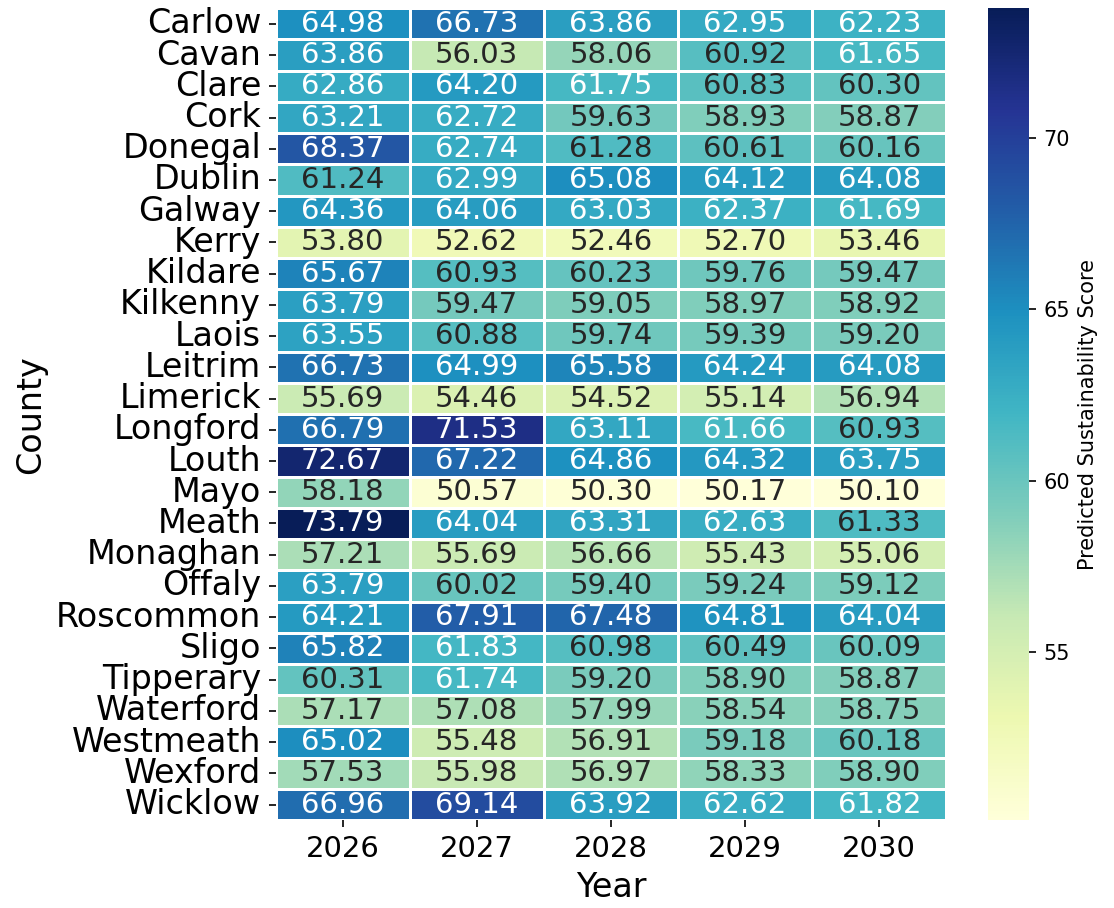}
        \caption{}
        \label{fig:sub1}
    \end{minipage}
    \hfill
    \begin{minipage}{0.45\textwidth}
        \centering
        \includegraphics[width=\linewidth]{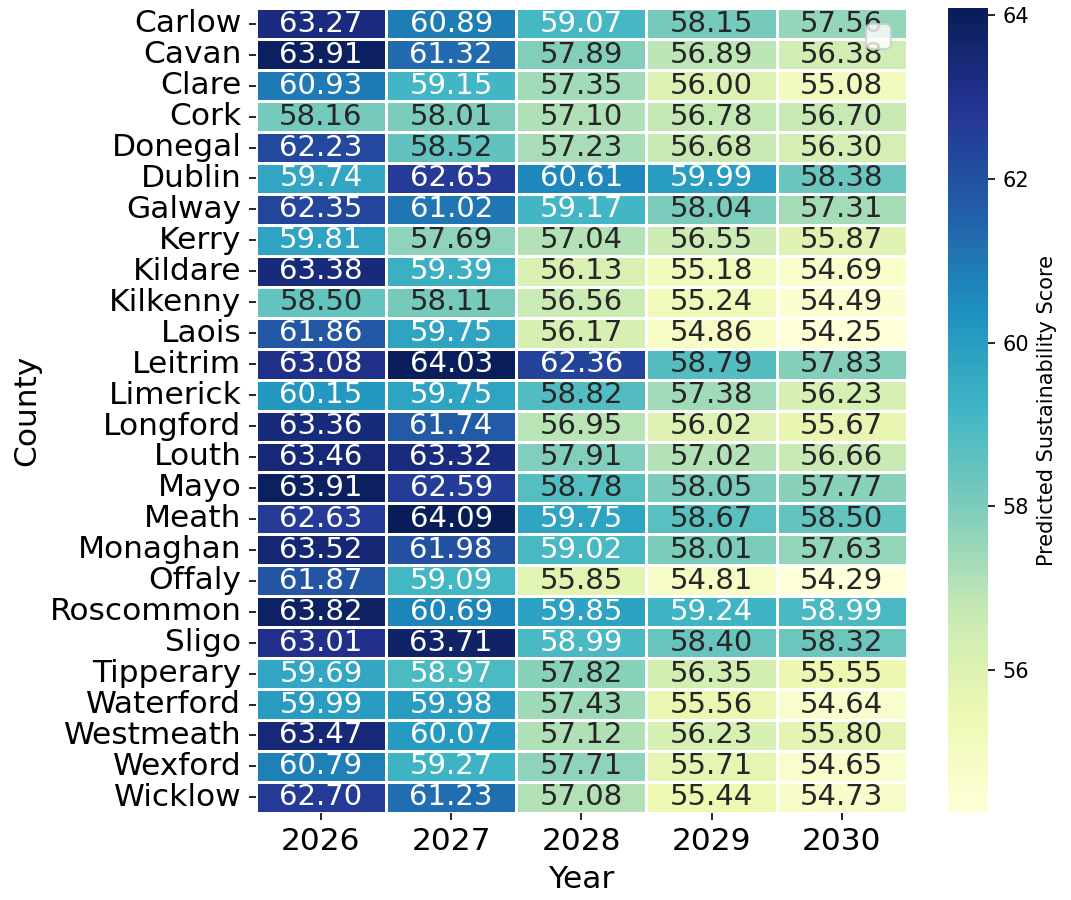}
        \caption{}
        \label{fig:sub2}
    \end{minipage}

    \vspace{0.3cm}

    \centering
    \begin{minipage}{0.45\textwidth}
        \centering
        \includegraphics[width=\linewidth]{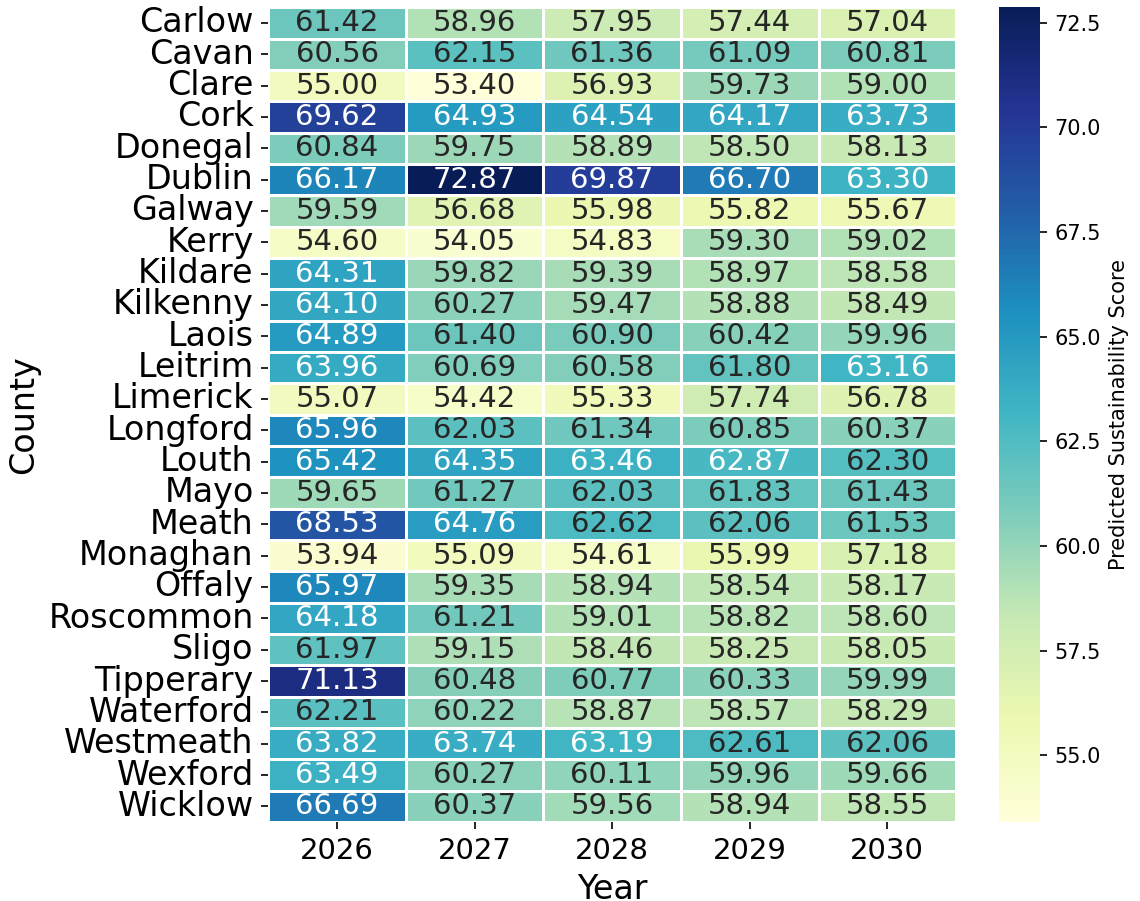}
        \caption{}
        \label{fig:sub1}
    \end{minipage}
    \hfill
    \begin{minipage}{0.45\textwidth}
        \centering
        \includegraphics[width=\linewidth]{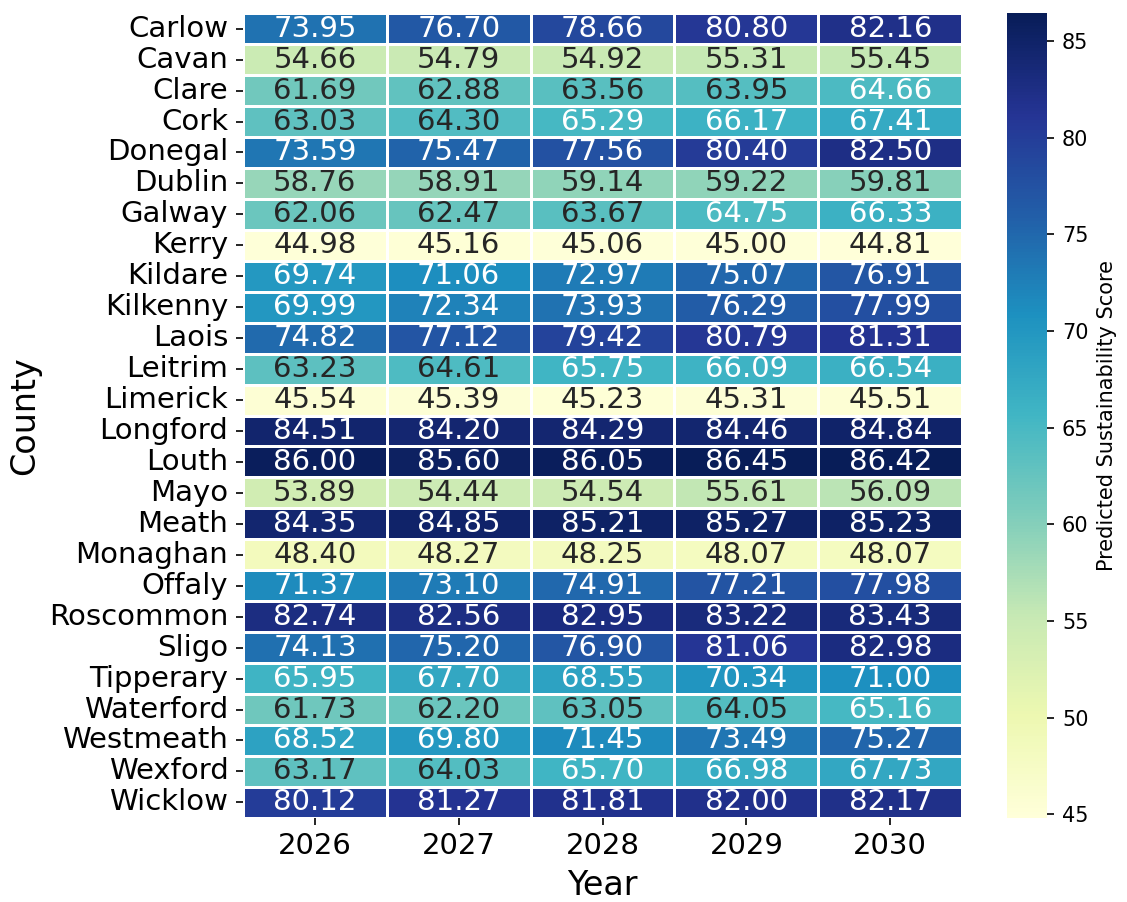}
        \caption{}
        \label{fig:sub2}
    \end{minipage}

    \caption{Heatmap comparison of baseline model forecasts (2026-2030): (a) RNN, (b) LSTM, (c) FFNN, (d) GKR.}
    \vspace{-6mm}
    \label{fig:combined}
\end{figure}

Crucially, the STGNN forecast remains spatially differentiated. High-performing counties at the end of the historical period remain leaders in the projection (e.g., Wicklow reaching 95.4 by 2030 and Longford 92.5 in Table~\ref{tab:county_trends}), while historically weaker counties improve but do not fully converge (e.g., Monaghan at 75.3 by 2030). This behavior is physically consistent with partial spatial coupling: neighboring counties share agro-climatic conditions, advisory networks, and market/infrastructure effects, but persistent differences in herd structure and management constraints prevent uniform convergence. The retained ranking also suggests that the model avoids over-smoothing across the county graph. The sustainability score is an aggregated proxy (derived via PCA from standardized herd-level indicators into four pillars) and therefore reflects changes in management-relevant drivers of environmental efficiency rather than direct emission measurements. Under this interpretation, the projected upward trends imply improving efficiency (e.g., reduced unproductive days and losses), which aligns with lower replacement demand and reduced resource use per unit output at county scale.

\begin{table}[t!]
\centering
\caption{Trends in Historical and Predicted Sustainability Scores for Selected Counties}
\begin{tabular}{l l p{1.2cm} p{1.2cm} p{1.2cm}}
\hline
\textbf{Trend Category} & \textbf{County} & \textbf{2025} & \textbf{2026} & \textbf{2030} \\
\hline
Highest Trend Leader & Wicklow  & 88.7 & 85.5 & 95.4 \\
                    & Longford & 92.2 & 86.6 & 92.5 \\
Lowest Trend Anchor & Monaghan & 50.1 & 51.7 & 75.3 \\
                    & Kerry    & 40.1 & 65.0 & 84.7 \\
                    & Leitrim  & 41.3 & 70.4 & 76.9 \\
\hline
\end{tabular}
\label{tab:county_trends}
\end{table}

Fig.~\ref{fig:combined} contrasts baseline model forecasts (RNN, LSTM, FFNN, GKR) with the STGNN behavior observed in Fig.~\ref{fig:sustainability score}. The baselines exhibit characteristic structural limitations. Sequence-only models (RNN/LSTM) tend to under-represent spatial heterogeneity and produce comparatively flattened forecasts, reflecting their inability to propagate information across counties. The FFNN, lacking explicit sequence modeling, yields overly uniform predictions and a compressed range, which is inconsistent with persistent regional differences. In contrast, GKR shows excessive dispersion and saturation effects, culminating in near-ceiling predictions for many counties by 2030, a pattern that is difficult to justify in real systems where biological and managerial improvements are gradual and bounded. Overall, the heatmap comparison supports the premise that credible sustainability forecasting at county scale requires \emph{both} temporal learning and explicit spatial coupling. The STGNN achieves this balance: it smooths short-term volatility while preserving regional ranking, producing stable and actionable medium-term projections.

\subsection{Uncertainty Quantification of County-Level Sustainability Forecasts}

\begin{figure}[t]
    \centering
    \includegraphics[width=\textwidth]{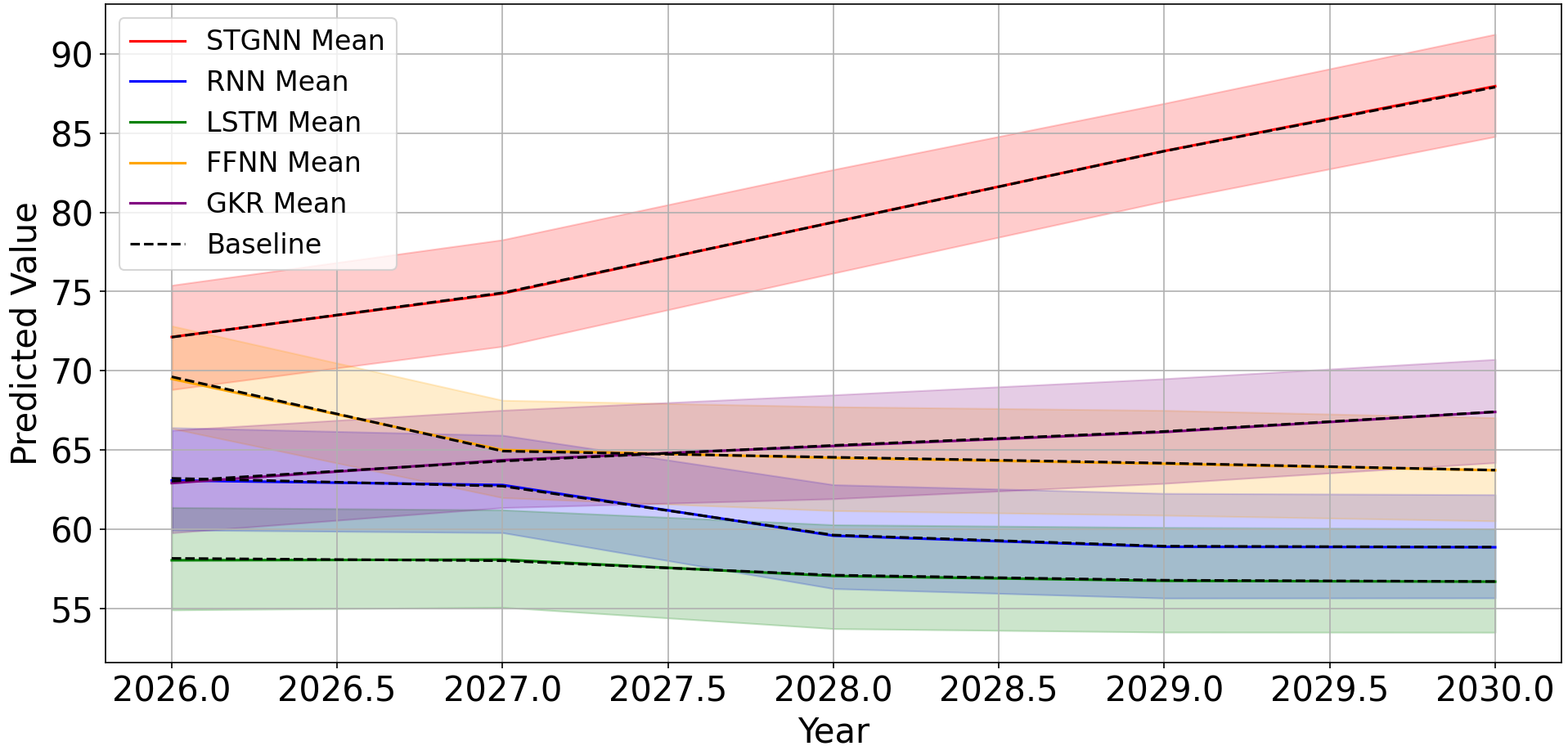}
    \caption{County sustainability score forecast with uncertainty for Cork (Monte Carlo, 2026-2030).}
    \label{fig:montecarloCork}
\end{figure}

\begin{figure}[t]
    \centering
    \includegraphics[width=\textwidth]{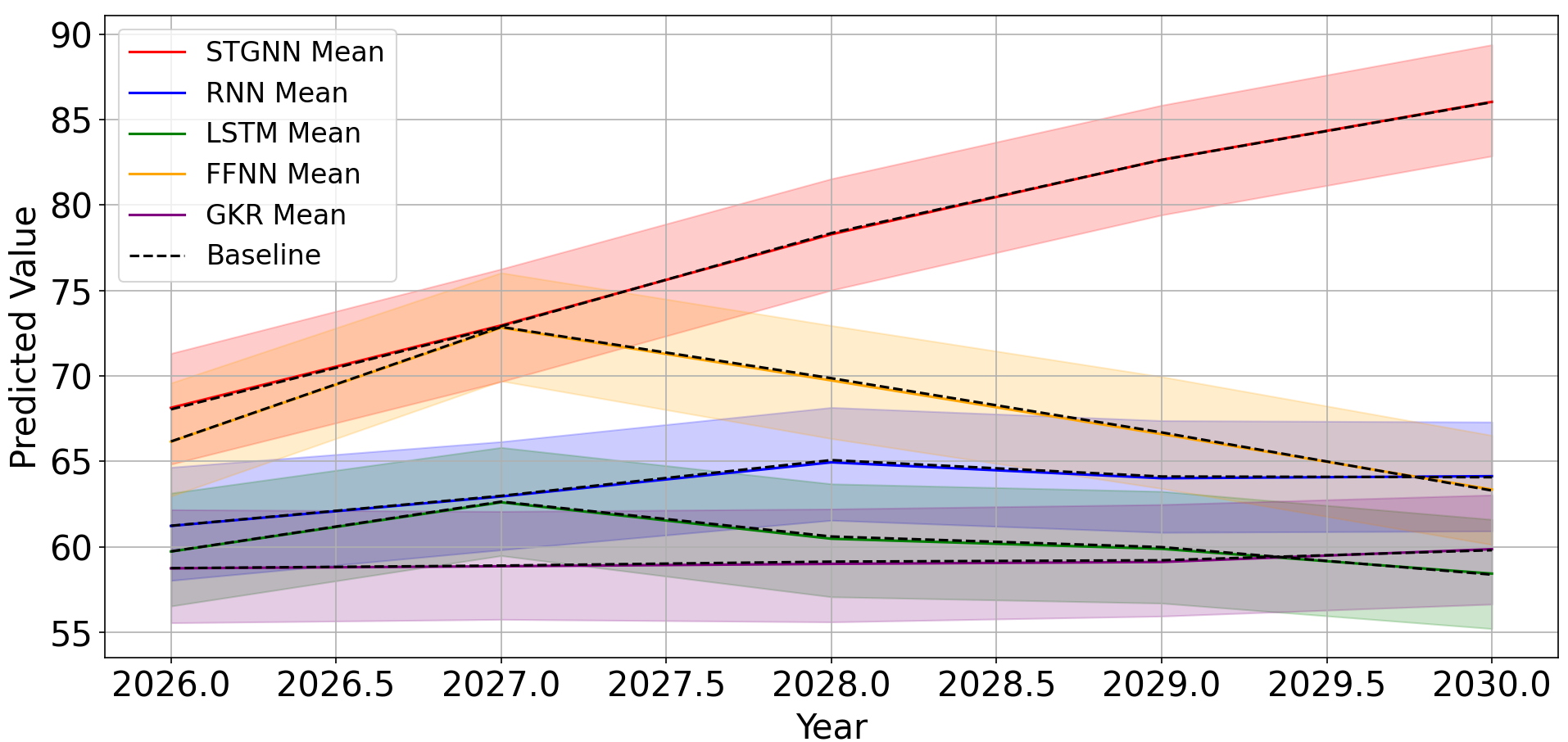}
    \caption{County sustainability score forecast with uncertainty for Dublin (Monte Carlo, 2026-2030).}
    \label{fig:montecarlodublin}
\end{figure}

To quantify forecast risk, Monte Carlo simulations were performed, and the mean trajectory, along with the 5–95\% confidence interval (CI), is reported for representative counties. Fig.~\ref{fig:montecarloCork} (Cork) and Fig.~\ref{fig:montecarlodublin} (Dublin) show that the STGNN produces the most reliable probabilistic forecasts among all models: it yields the highest mean trajectory and the tightest CI across 2026-2030, indicating low sensitivity to stochastic perturbations and data resampling. In both counties, the STGNN predicts a sustained improvement (Cork $\approx 72\!\rightarrow\!88$, Dublin $\approx 72\!\rightarrow\!87$), consistent with gradual, cumulative changes in management-driven indicators rather than abrupt year-to-year swings.

The CI bands widen modestly with forecast horizon, which is expected in forward prediction as unmodeled variability accumulates over time. For Cork, the CI remains narrow in early years (high short-term certainty) and expands by 2030 (approximately mid-80s to low-90s), providing stakeholders with a practical range of plausible outcomes. Physically, this widening reflects compounding uncertainty in the operational drivers aggregated into the sustainability score (e.g., fertility performance, replacement pressure, and loss events), which are influenced by weather shocks, disease incidence, and heterogeneous adoption of interventions.

In contrast, the baseline models exhibit either \emph{trend attenuation} or \emph{instability}. RNN and LSTM typically produce flatter mean paths and underestimate improvements, reflecting their inability to leverage inter-county coupling. FFNN collapses toward a low, nearly static baseline, consistent with underfitting when temporal structure is not modeled. GKR shows exaggerated volatility and saturation behavior (occasionally approaching ceiling values), yielding wide CIs and biologically implausible dynamics for a management-driven system where improvements are bounded and incremental. The limited overlap between the STGNN CI and those of the weaker baselines (notably FFNN and sequence-only models) further supports that the STGNN’s advantage is not only in point accuracy but also in uncertainty-aware robustness, which is essential for policy-relevant, medium-term planning.

\subsection{Impact of Targeted Interventions on Sustainability: Counterfactual Analysis}

\begin{figure}[t]
    \centering

    \begin{minipage}[t]{0.48\textwidth}
        \centering
        \textbf{Monaghan}\par\vspace{0.6em}

        \includegraphics[width=\linewidth]{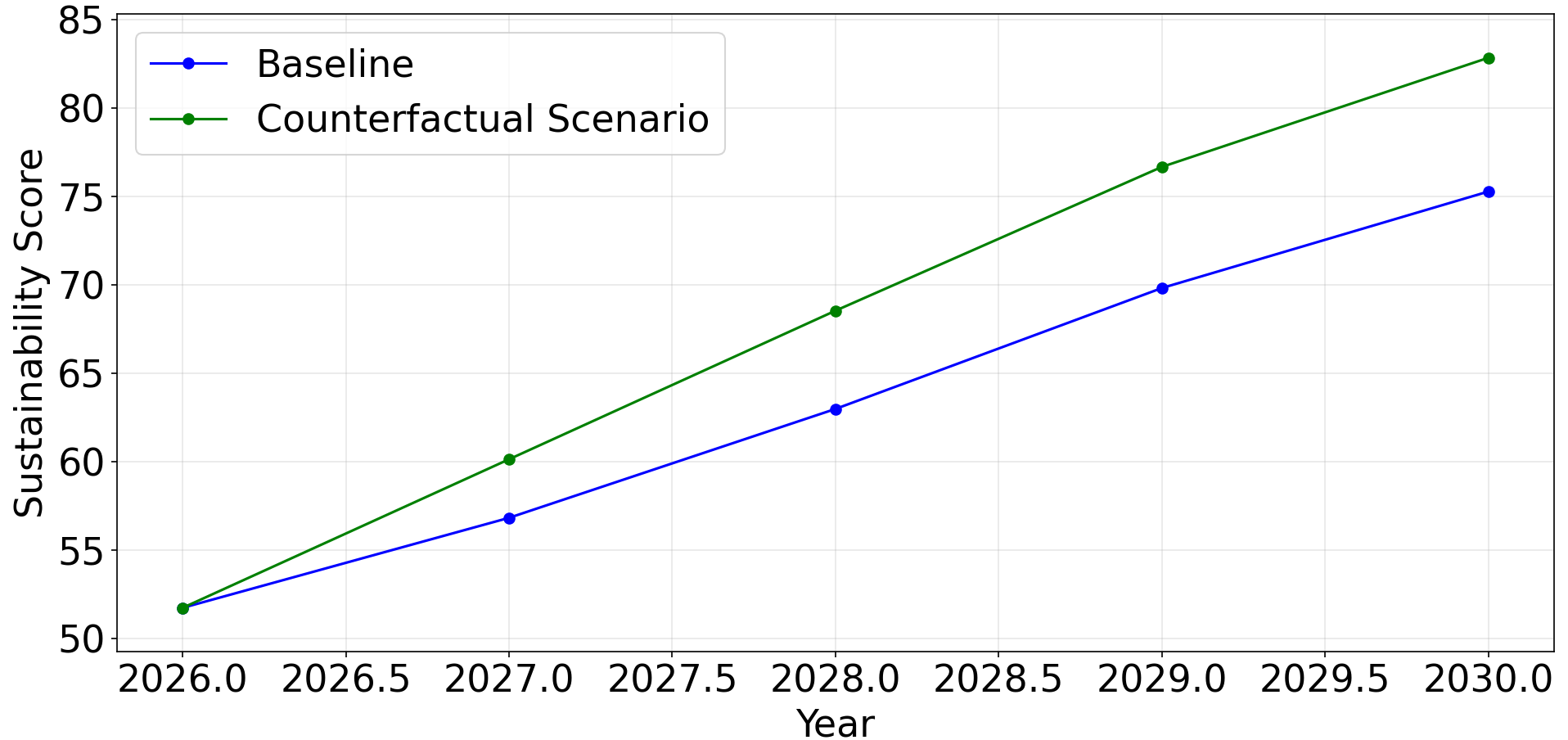}
        \par\small (a) STGNN\par\vspace{0.8em}

        \includegraphics[width=\linewidth]{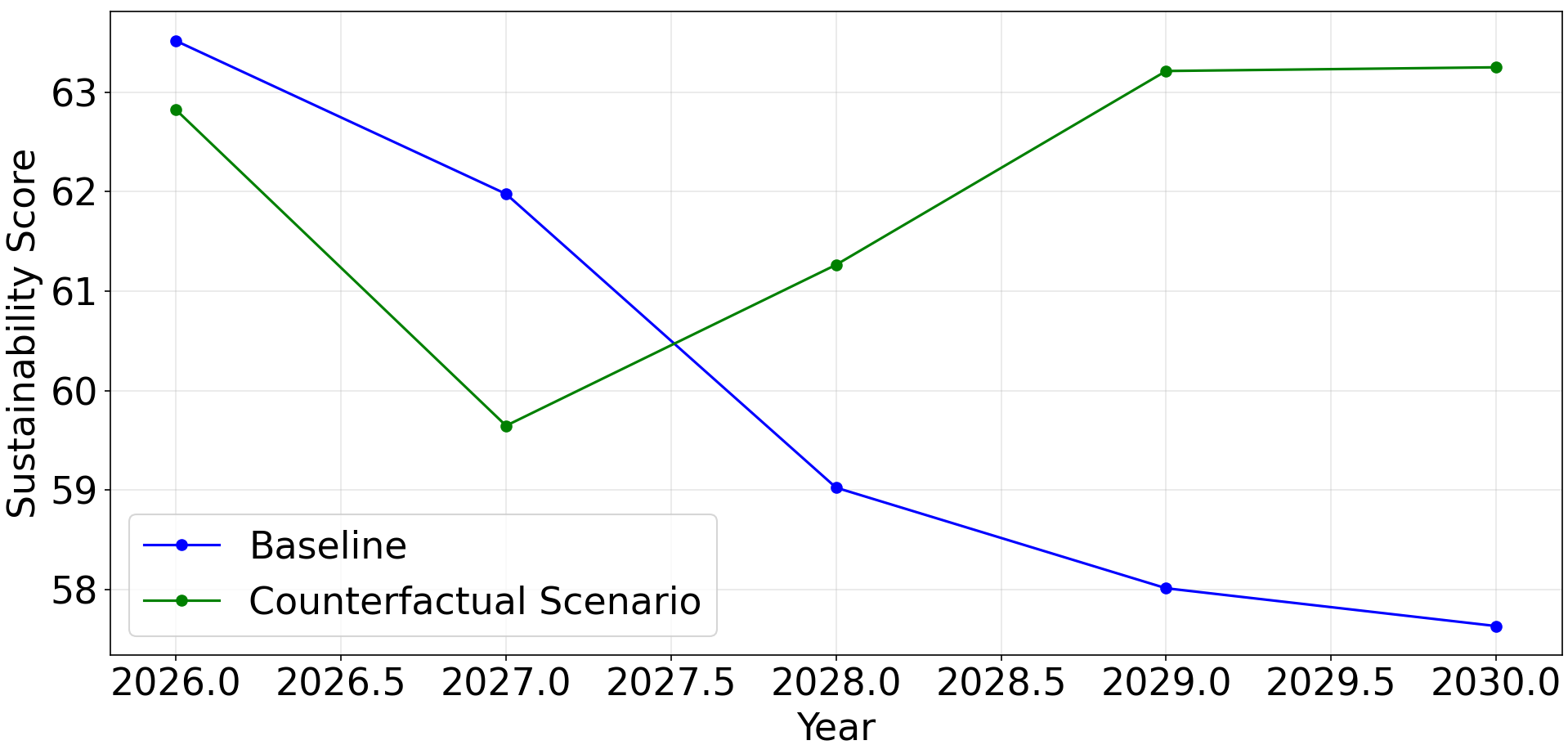}
        \par\small (b) LSTM\par\vspace{0.8em}

        \includegraphics[width=\linewidth]{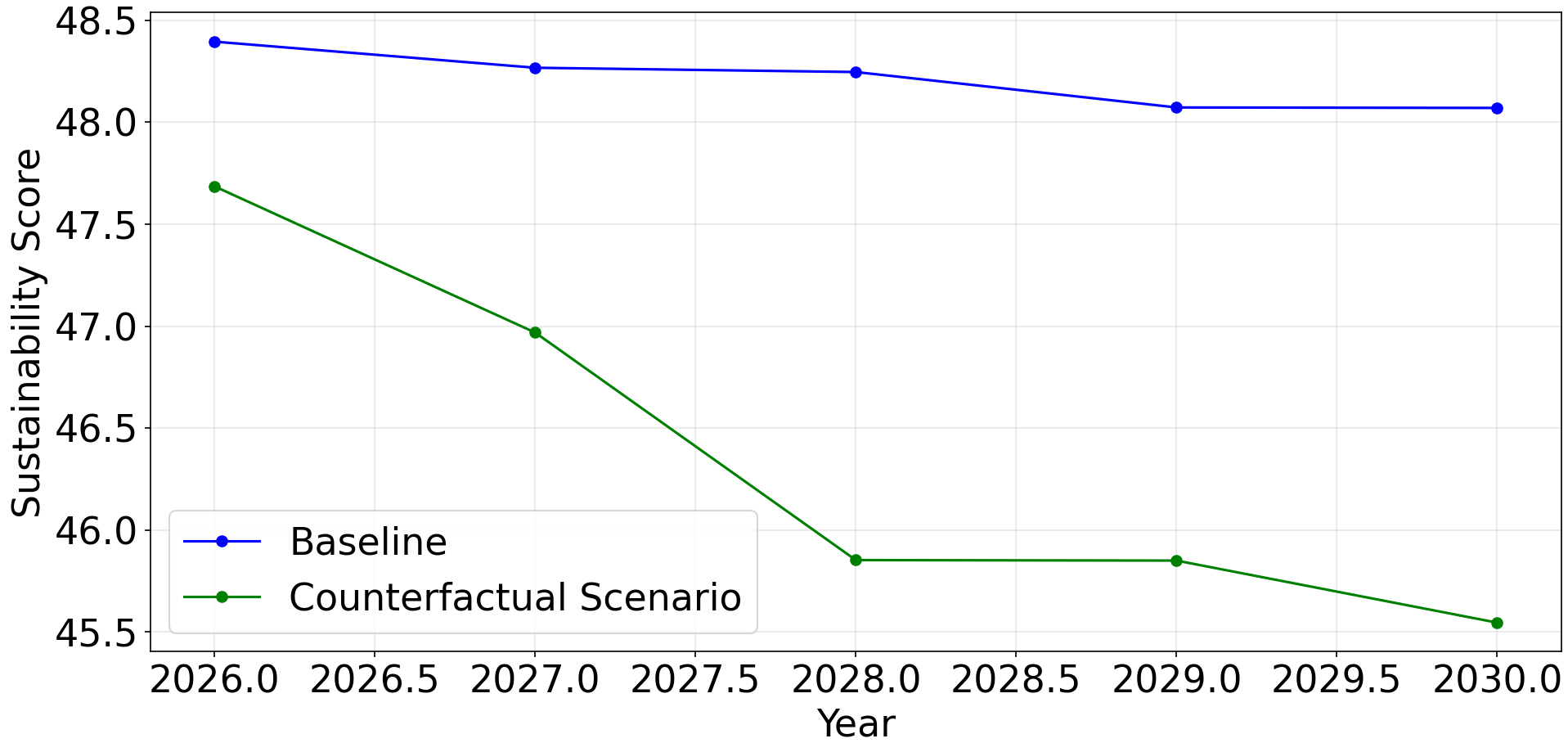}
        \par\small (c) GKR
    \end{minipage}
    \hfill
    \begin{minipage}[t]{0.48\textwidth}
        \centering
        \textbf{Kerry}\par\vspace{0.6em}

        \includegraphics[width=\linewidth]{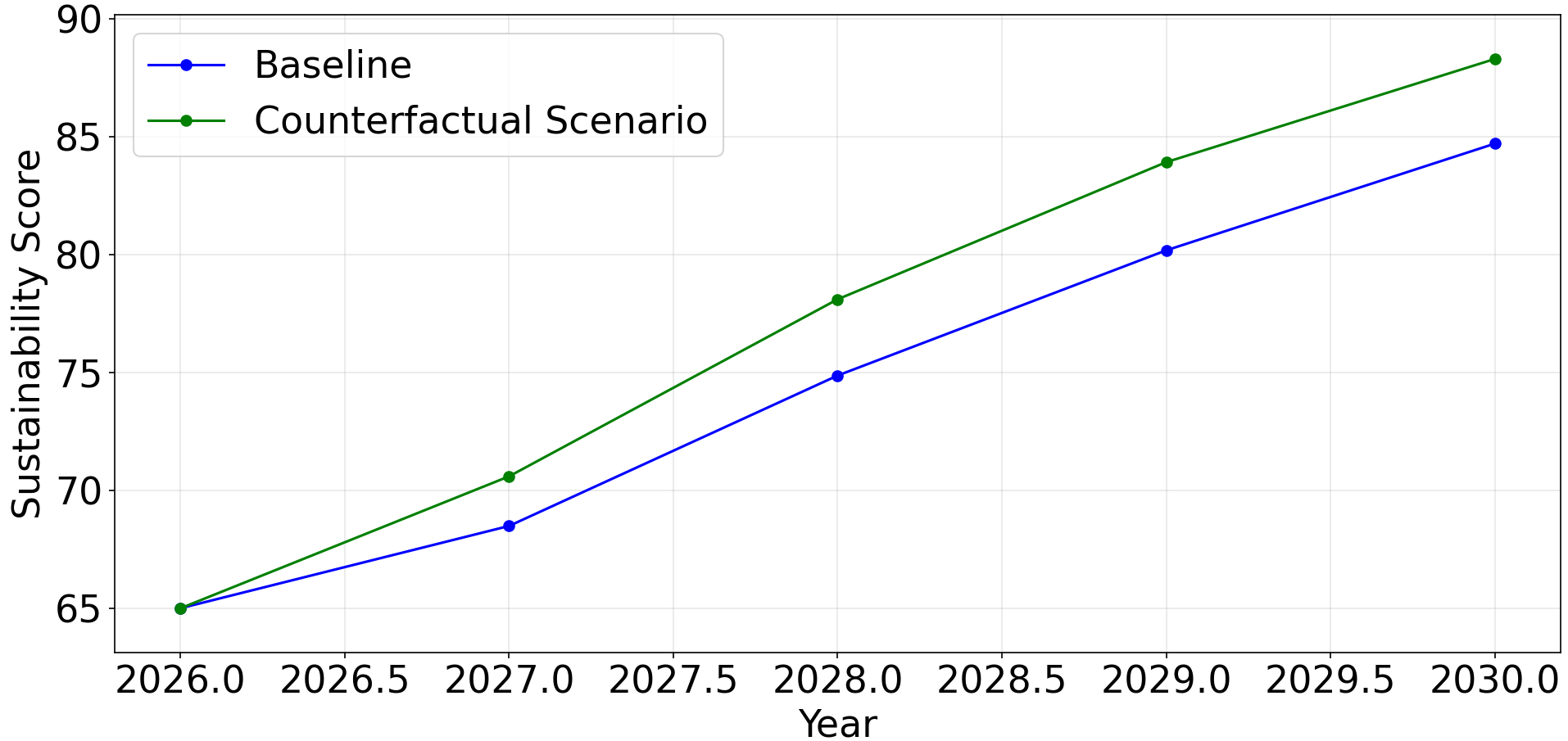}
        \par\small (d) STGNN\par\vspace{0.8em}

        \includegraphics[width=\linewidth]{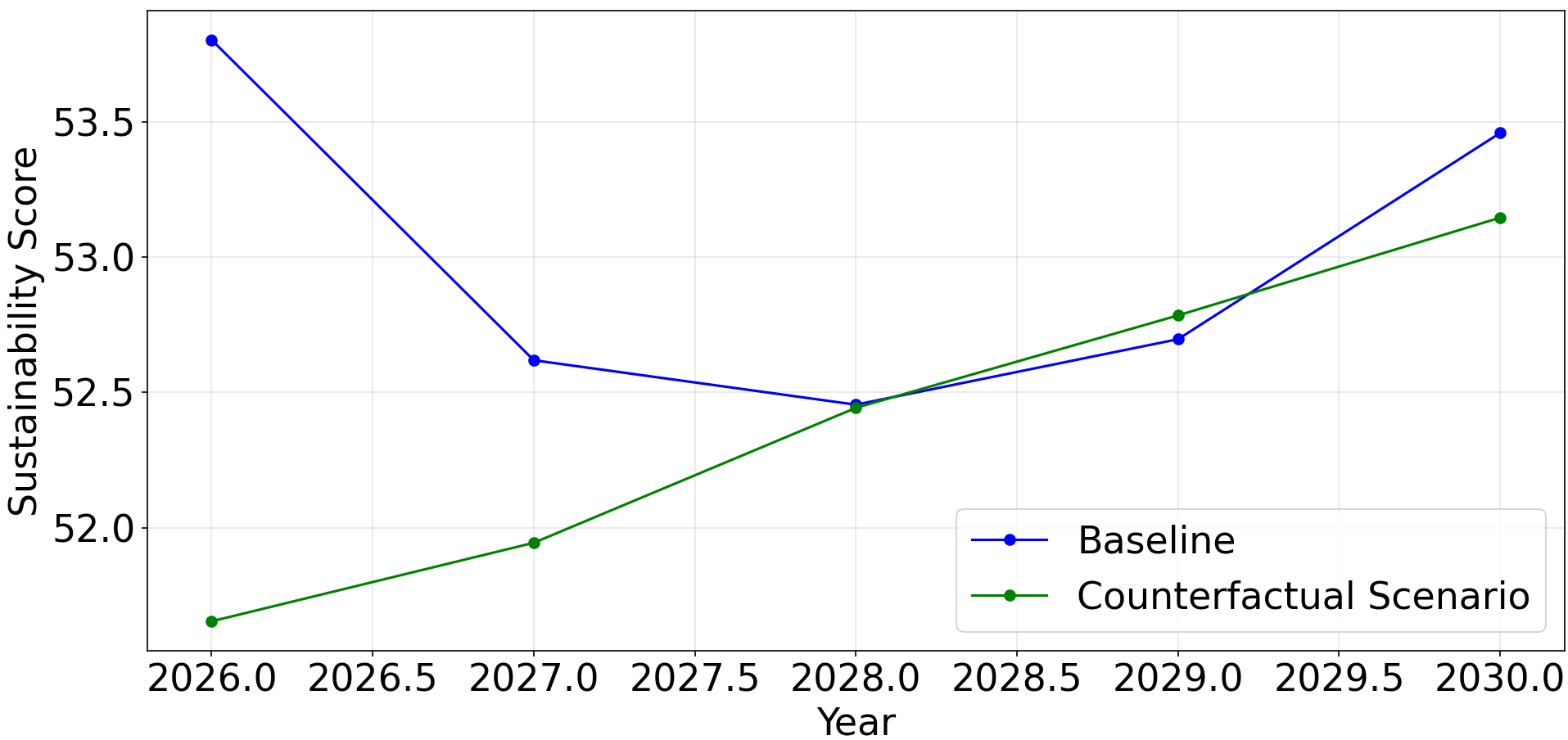}
        \par\small (e) RNN\par\vspace{0.8em}

        \includegraphics[width=\linewidth]{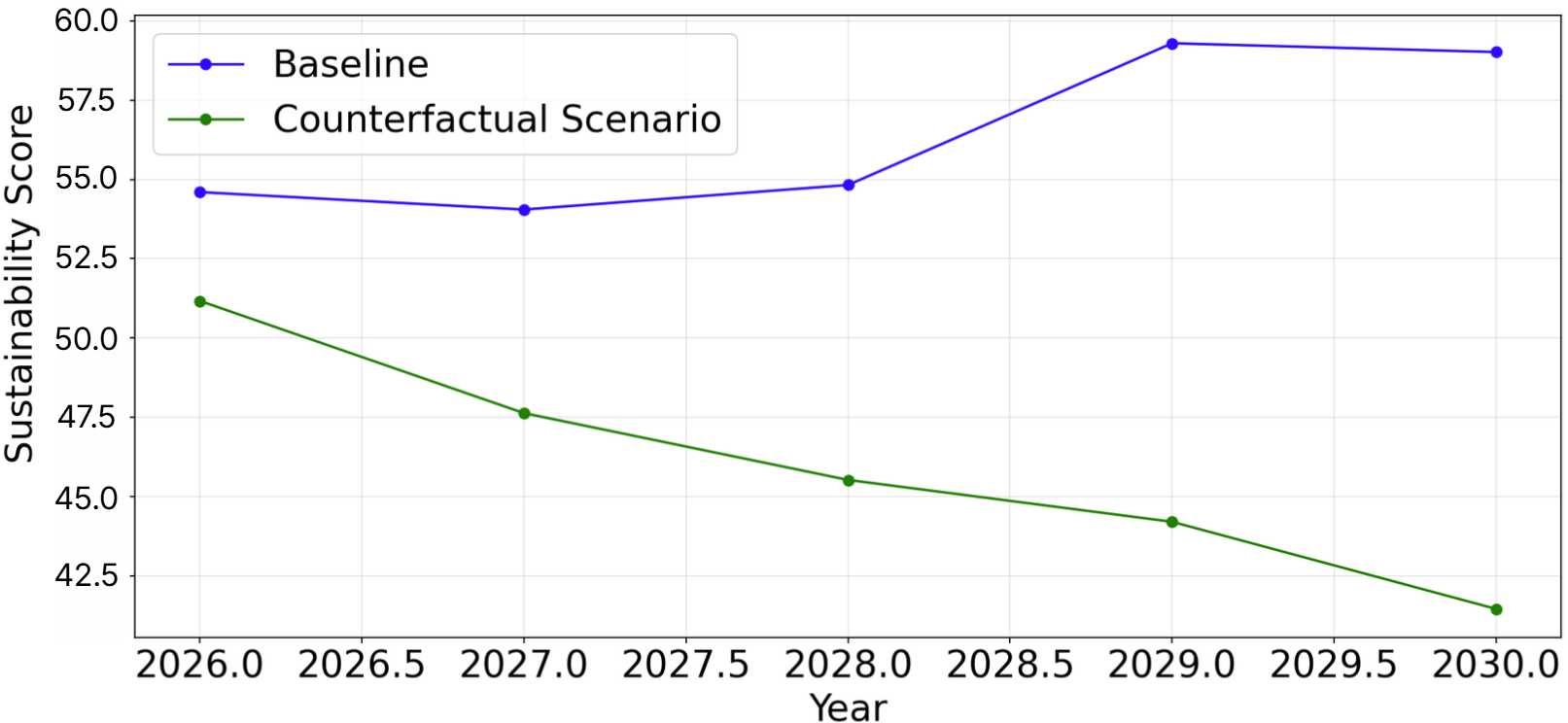}
        \par\small (f) FFNN
    \end{minipage}

    \caption{Counterfactual forecast comparison (2026--2030) for two counties.
    (a--c) Monaghan: STGNN, LSTM, and GKR.
    (d--f) Kerry: STGNN, RNN, and FFNN.}
    \label{fig:counterfactual}
\end{figure}

Counterfactual experiments evaluate how targeted improvements in high-impact operational indicators translate into future county sustainability trajectories. Fig.~\ref{fig:counterfactual} compares the baseline forecast (continuation of current trends) against a ``what-if'' intervention that perturbs selected high-loading features linked to the PCA-derived pillars (notably Reproductive Efficiency and Herd Health). Because the sustainability score is an operational proxy, the counterfactual lift should be interpreted as expected gains in management efficiency that are known to reduce replacement pressure, avoidable losses, and resource use per unit output.

\textbf{Monaghan:} Fig.~\ref{fig:counterfactual}(a) shows that STGNN converts the intervention into a sustained and widening improvement relative to baseline, increasing the 2030 score from $\approx 75.3$ to $\approx 82.5$ (a gain of $\approx 7.2$ points). This response is physically plausible: improvements such as shorter calving interval and reduced culling yield compounding benefits over multiple production cycles, lowering the proportion of nonproductive animals and stabilizing herd structure. In contrast, Fig.~\ref{fig:counterfactual}(b)-(c) illustrate that LSTM and GKR do not reproduce this compounding effect reliably; the predicted gaps are smaller and/or exhibit non-monotonic behavior, indicating limited ability to propagate intervention effects through coupled spatial-temporal dynamics.

\textbf{Kerry:} A similar pattern is observed in Fig.~\ref{fig:counterfactual}(d-f). The STGNN counterfactual trajectory remains consistently above the baseline and reaches $\approx 88.5$ by 2030, compared with $\approx 84.7$ under baseline (a gain of $\approx 3.8$ points). The sustained separation indicates that the model treats the intervention as a persistent shift in underlying drivers rather than a transient perturbation. Baseline models (e.g., RNN and FFNN in Fig.~\ref{fig:counterfactual}(e)-(f)) show muted, unstable, or poorly scaled responses, which is consistent with their inability to jointly represent (i) gradual temporal accumulation and (ii) spatial spillovers across neighboring counties.

Overall, the counterfactual results highlight two policy-relevant properties. First, lower-performing counties exhibit measurable headroom: modest changes in a small set of high-leverage indicators produce persistent gains over the 5-year horizon. Second, credible intervention simulation requires explicit spatio-temporal structure; only the STGNN consistently yields monotone, bounded, and interpretable improvements that align with the gradual nature of biological and management responses in dairy systems.

\section{Conclusion}

In this paper, a novel data-driven framework for county-scale assessment and forecasting of dairy sustainability in Ireland, using herd-level operational records, is presented. To the best of current knowledge, this study provides a \emph{first-ever} integrated pipeline that combines VAE-based data augmentation, PCA-derived farm-level pillars (Reproductive Efficiency, Genetic Management, Herd Health, and Herd Management), and an attention-enhanced STGNN to jointly learn temporal dynamics and spatial coupling across counties. The composite sustainability score offers an interpretable and stable basis for benchmarking and medium-term forecasting. This study further strengthened decision relevance by quantifying uncertainty through Monte Carlo simulation and by performing counterfactual analyses that translate targeted changes in high-leverage indicators into measurable improvements in future trajectories, highlighting actionable headroom for lower-performing regions. Although the score is proxy-based rather than a direct emissions measurement, it supports a clear physical interpretation: improvements in fertility, health, loss reduction, and breeding efficiency reduce replacement pressure and resource use per unit output, thereby lowering waste and emissions intensity. Overall, the results indicate continued county-level improvements through 2030 and it is demonstrated that the spatio–temporal approach yields robust, policy-relevant forecasts and intervention-sensitive projections.

The future work will extend this framework in four directions. First,  will incorporate \emph{direct} environmental and climate variables including GHG emissions and emissions intensity, nutrient balances and runoff risk, water and energy use, and soil-health indicators, to move from operational proxies toward explicit environmental attribution and validation. Second, will expand to a \emph{full multi-pillar} sustainability evaluation by integrating economic viability (e.g., profitability, input-cost exposure, market volatility) and social/welfare dimensions (e.g., welfare outcomes, antimicrobial-use proxies, labor and resilience indicators), enabling systematic trade-off analysis. Third,  will generalize scenario modeling beyond local parameter shifts by embedding policy, market, and climate pathways (e.g., regulatory constraints, feed-price shocks, heat-stress and drought regimes) to support long-horizon planning under deep uncertainty. Fourth,  will strengthen scalability and transferability by applying the approach perfomed in this study to other livestock sectors and regions, and by studying improved graph construction (e.g., mobility, proximity, and supply-chain-informed graphs), domain adaptation, and causal counterfactual formulations to enhance interpretability and intervention credibility. These extensions will position the framework as a comprehensive, transferable decision-support tool for data-driven sustainable livestock management.

\noindent\textbf{Acknowledgements} This study was supported by the Department of Agriculture, Food and the Marine (DAFM) through the 2023 Thematic Research Call, under which the AgNav project was funded. The project builds on a collaborative partnership between Teagasc, the Irish Cattle Breeding Federation (ICBF), and Bord Bia, who together developed the AgNav digital platform.

\noindent\textbf{Funding} This research has emanated from research conducted with the financial support of DAFM 2023 Thematic Research Call. The project reference number is 2023RP956.



\begin{thebibliography}{00}

\bibitem[1]{Mohammadzadeh2025}
Mohammadzadeh, M., Hayati, D., \& Valizadeh, N. (2025).
Behavioral factors linking sustainability and animal welfare in dairy farming.
\textit{Scientific Reports, 15}, 26042.
https://doi.org/10.1038/s41598-025-10260-2

\bibitem[2]{Xu2023}
Xu, S., Li, X., Zhang, Y., \& Fan, S. (2023).
Spatiotemporal dynamics of agricultural sustainability assessment: A study across 30 Chinese provinces.
\textit{Sustainability, 15}(11), 9066.
https://doi.org/10.3390/su15119066

\bibitem[3]{AHI2024}
Animal Health Ireland. (2024).
\textit{Sustainable agriculture}.
https://animalhealthireland.ie/about/sustainable-agriculture/

\bibitem[4]{deOliveira2024}
de Oliveira, F. M., Ferraz, G. A. E. S., André, A. L. G., Santana, L. S., Norton, T., \& Ferraz, P. F. P. (2024).
Digital and precision technologies in dairy cattle farming: A bibliometric analysis.
\textit{Animals, 14}(12), 1832.
https://doi.org/10.3390/ani14121832


\bibitem[5]{Shine2022}
Shine, P., \& Murphy, M. D. (2022).
Over 20 years of machine learning applications on dairy farms: A comprehensive mapping study.
\textit{Sensors, 22}(1), 52.
https://doi.org/10.3390/s22010052

\bibitem[6]{Sommerseth2024}
Sommerseth, J. K., Shrestha, S., MacLeod, M., Hegrenes, A., Hansen, B. G., \& Salte, R. (2024).
How increased heifer growth rate and reduced dairy cow replacement rate can improve farm economy and reduce greenhouse gas emissions: A win-win situation?
\textit{Animal, 18}(9), 101294.
https://doi.org/10.1016/j.animal.2024.101294

\bibitem[7]{Crowe2018}
Crowe, M. A., Hostens, M., \& Opsomer, G. (2018).
Reproductive management in dairy cows – The future.
\textit{Irish Veterinary Journal, 71}(1), 1.
https://doi.org/10.1186/s13620-017-0112-y

\bibitem[8]{Rios2020}
Rios, G. P., \& Botero, S. (2020).
An integrated indicator to analyze sustainability in specialized dairy farms in Antioquia-Colombia.

\textit{Sustainability, 12}(22), 9595.
https://doi.org/10.3390/su12229595


\bibitem[9]{Diaz2021}
D{\'i}az de Ot{\'a}lora, X., del Prado, A., Dragoni, F., Estell{\'e}s, F., \& Amon, B. (2021).
Evaluating three-pillar sustainability modelling approaches for dairy cattle production systems.
\textit{Sustainability, 13}(11), 6332.
https://doi.org/10.3390/su13116332

\bibitem[10]{Arvidsson2020}
Arvidsson Segerkvist, K., Hansson, H., Sonesson, U., \& Gunnarsson, S. (2020).
Research on environmental, economic, and social sustainability in dairy farming: A systematic mapping of current literature.
\textit{Sustainability, 12}(14), 5502.
https://doi.org/10.3390/su12145502

\bibitem[11]{VanEenennaam2025}
Van Eenennaam, A. L. (2025).
Current and future uses of genetic improvement technologies in livestock breeding programs.
\textit{Animal Frontiers, 15}(1), 80–90.
https://doi.org/10.1093/af/vfae042



\bibitem[12]{Diavao2023}
Diav{\~a}o, J., et al. (2023).
How does reproduction account for dairy farm sustainability?
\textit{Animal Reproduction, 20}(2), e20230066.
https://doi.org/10.1590/1984-3143-AR2023-0066


\bibitem[13]{Crowe2018}
Crowe, M. A., Hostens, M., \& Opsomer, G. (2018).
Reproductive management in dairy cows – the future.
\textit{Irish Veterinary Journal, 71}(1), 1.
https://doi.org/10.1186/s13620-017-0112-y



\bibitem[14]{deHaas2023}
de Haas, Y., Pryce, J. E., Berry, D. P., \& Veerkamp, R. F. (2023).
Genetic and genomic solutions to improve feed efficiency and reduce environmental impact of dairy cattle.
\textit{Animal Breeding \& Genomics Conference Proceedings}.

\bibitem[15]{Muruganantham2022}
Muruganantham, P., Wibowo, S., Grandhi, S., Samrat, N. H., \& Islam, N. (2022).
A systematic literature review on crop yield prediction with deep learning and remote sensing.
\textit{Remote Sensing, 14}(9), 1990.
https://doi.org/10.3390/rs14091990

\bibitem[16]{Gupta2023}
Gupta, A., \& Singh, A. (2023).
Agri-GNN: A novel genotypic-topological graph neural network framework built on GraphSAGE for optimized yield prediction.
\textit{arXiv}.
https://arxiv.org/abs/2310.13037


\bibitem[17]{Vyas2020}
Vyas, A., \& Bandyopadhyay, S. (2020).
Dynamic structure learning through graph neural network for forecasting soil moisture in precision agriculture.
In \textit{Proceedings of the 29th International Joint Conference on Artificial Intelligence} (pp. 720-726).
https://arxiv.org/abs/2012.03506


\bibitem[18]{Sophocleous2024}
Sophocleous, M., Karkotis, A., Papasavva, A., et al. (2024).
A stand-alone, in situ, soil quality sensing system for precision agriculture.
\textit{IEEE Transactions on AgriFood Electronics, 2}(1), 43-50.
https://doi.org/10.1109/TAFE.2024.3351953


\bibitem[19]{ICBF2025}
Irish Cattle Breeding Federation. (2025).
\textit{ICBF dairy statistics web application}.
https://webapp.icbf.com/v2/app/dairy-stats

\bibitem[20]{Clasen2024}
Clasen, J.B., Fikse, W.F., Ramin, M., Lindberg, M., et al. (2024). 
Effects of herd management decisions on dairy cow longevity, farm profitability, and emissions of enteric methane – a simulation study of milk and beef production. 
\textit{Animal, 18}(2), 101051. 
https://doi.org/10.1016/j.animal.2023.101051



\end{thebibliography}
\end{document}